\documentclass[conference]{IEEEtran}
\IEEEoverridecommandlockouts

\usepackage{cite}
\usepackage{amsmath,amssymb,amsfonts}
\usepackage{graphicx}
\usepackage{textcomp}
\usepackage{xcolor}

\usepackage{bm}
\usepackage[dvipsnames]{xcolor}
\usepackage{multirow}
\usepackage{amsmath}
\usepackage{amssymb}
\usepackage{mathtools}
\usepackage[inline]{enumitem}
\usepackage{graphicx}
\usepackage{booktabs}
\usepackage{tabularx}
\usepackage{pifont}
\usepackage{hyperref}
\usepackage{mdframed}
\usepackage{dashrule}
\usepackage[caption=false, font=footnotesize]{subfig}
\usepackage{diagbox}
\usepackage{xspace}
\usepackage{spverbatim}
\usepackage[figuresright]{rotating}
\usepackage{scalerel}

\usepackage{algorithm,algpseudocode}

\algnewcommand{\LineComment}[1]{\State \(\triangleright\) #1}

\DeclarePairedDelimiterX{\infdivx}[2]{(}{)}{%
  #1\;\delimsize\|\;#2%
}

\newcommand{\sciagent}{SciDER\xspace}
\newcommand{\opensci}{OpenSciDER\xspace}

\newcommand{\gmark}{\textcolor{ForestGreen}{\ding{52}}}
\newcommand{\rmark}{\textcolor{BrickRed}{\ding{56}}}
\newcommand{\cxmark}{\ding{51}\hspace{-1.75mm}\ding{55}}
\newcommand{\ymark}{\textcolor{Dandelion}{\cxmark}}

\NewDocumentCommand{\kelin}{ mO{} }{\textcolor{cyan}{\textsuperscript{\textit{KeLin}}\textsf{\textbf{\small[#1]}}}}

\def\BibTeX{{\rm B\kern-.05em{\sc i\kern-.025em b}\kern-.08em
    T\kern-.1667em\lower.7ex\hbox{E}\kern-.125emX}}
\begin{document}

\title{\sciagent: Scientific Data-centric End-to-end Researcher
\thanks{*The project is available at \url{https://github.com/leonardodalinky/SciDER}.}
}

\author{\IEEEauthorblockN{Ke Lin}
\IEEEauthorblockA{\textit{Data Science} \\
\textit{College of William \& Mary}\\
Williamsburg, USA \\
leonard.keilin@gmail.com}
\and
\IEEEauthorblockN{Owais Aijaz}
\IEEEauthorblockA{\textit{Natural Language Processing} \\
\textit{MBZUAI}\\
Abu Dhabi, UAE \\
owais.aijaz@mbzuai.ac.ae}
\and
\IEEEauthorblockN{Yilin Lu}
\IEEEauthorblockA{\textit{Computer Science} \\
\textit{University of Minnesota}\\
Minneapolis, USA \\
lu000661@umn.edu}
\and
\IEEEauthorblockN{Yiyang Luo}
\IEEEauthorblockA{\textit{Academy of Interdisciplinary Studies} \\
\textit{Hong Kong University of Science and Technology}\\
Hong Kong, China \\
yluodq@connect.ust.hk}
\and
\IEEEauthorblockN{Xuehang Guo}
\IEEEauthorblockA{\textit{Data Science} \\
\textit{College of William \& Mary}\\
Williamsburg, USA \\
xguo15@wm.edu}
\and
\IEEEauthorblockN{Preslav Nakov}
\IEEEauthorblockA{\textit{Natural Language Processing} \\
\textit{MBZUAI}\\
Abu Dhabi, UAE \\
preslav.nakov@mbzuai.ac.ae}
}

\maketitle

\begin{abstract}
While large language models accelerate scientific discovery, existing agents face severe limitations in adaptability, domain generalization, and multimodal scalability, often struggling to autonomously process raw, domain-specific experimental data. 
To overcome these barriers, we introduce \sciagent, a multi-agent system designed to flexibly automate the entire research lifecycle. 
This framework employs a novel data-centric approach and integrates a dynamic multimodal skill system across four specialized sub-agents. 
Specifically, an ideation agent generates novel hypotheses via Evolutionary Idea Search, a data analysis agent systematically structures raw data, an experimentation agent synthesizes executable code grounded in dataset characteristics, and a critic agent drives iterative self-refinement.
To democratize open-source scientific discovery, we release \opensci-SFT-8K, a high-quality execution trajectory dataset, alongside the \opensci-27B fine-tuned model. 
Across six benchmarks, SciDER and OpenSciDER obtain competitive or leading results, with especially strong gains on data-centric analysis, end-to-end research execution, and multimodal scientific visualization.
By integrating data analysis with experimental execution, \sciagent bridges the gap between abstract scientific reasoning and reproducible experimentation synthesis.
\end{abstract}

\begin{IEEEkeywords}
Autonomous agents, Scientific discovery, Data-centric AI, Multimodal systems, Large language models.
\end{IEEEkeywords}

\section{Introduction}

Large language models (LLM)-based agents are catalyzing scientific discovery~\cite{Wang2024GeneAgentSL,garikaparthi-etal-2025-iris,gottweis2025towards}. They can automate many research steps, from generating hypotheses to designing experiments, thereby speeding innovation and reducing barriers in scientific research. For example, AI Scientist v2~\cite{yamada2025ai} can autonomously write workshop-level papers. 
Despite this, creating a fully independent AI research partner still faces three unique barriers. 
The first challenge is the \textit{limited adaptability}. Most current systems are designed mainly for public machine learning datasets~\cite{chan2025mle, starace2025paperbenchevaluatingaisability, lu2024ai, tang2025ai}. Therefore, they struggle to independently analyze and process diverse real-world experimental data~\cite{zhang2025deepanalyze, luo2026benchmarkingaiscientistsomics}. 
The second challenge is the \textit{domain gap}. General-purpose assistants often fail in specialized domains where abstract ideas must be translated into precise, and sometimes proprietary, experimental data formats~\cite{white2023large,tian2024scicode,luo2026benchmarkingaiscientistsomics}.
The third challenge is the \textit{lack of multimodal scalability.} Previous works focus on unimodal workflows, limiting their ability to scale for multimodal tasks or leverage rapidly evolving AI skill plugins.

Fundamentally, these limitations stem from a decoupled approach to automated research. Traditional agentic frameworks often operate in a "top-down" manner, generating abstract hypotheses without systematically inspecting the underlying raw data first~\cite{tang2025ai, lu2024ai}. 
In contrast, real-world scientific discovery requires a highly iterative, data-driven workflow where researchers continuously process multimodal inputs, identify structural anomalies, and refine their experimental design accordingly. 
Without grounding the ideation and coding phases in autonomous, multi-perspective data analysis, existing systems inevitably struggle to bridge the gap between high-level reasoning and domain-specific execution~\cite{tian2024scicode, luo2026benchmarkingaiscientistsomics}.

\begin{table*}[tbp]
\centering
\newcolumntype{C}{>{\centering\arraybackslash}X}
\caption{
Comparison of research agent frameworks.
}

\begin{tabularx}{\textwidth}{@{}Cccccccccccc@{}}
\toprule
\multirow{2}{*}{\textbf{Framework}} & \multicolumn{4}{c}{\textbf{Functionality}} & \multicolumn{2}{c}{\textbf{Capability}} & \multicolumn{3}{c}{\textbf{Deployment}} & \multicolumn{2}{c}{\textbf{Open Source}}\\
\cmidrule(lr){2-5} \cmidrule(lr){6-7} \cmidrule(lr){8-10} \cmidrule(l){11-12}
 & \textbf{Ideation} & \textbf{Data} & \textbf{Experiment} & \textbf{Paper} & \textbf{Vision} & \textbf{Skill} & \textbf{Modular} & \textbf{Web UI} & \textbf{Package} & \textbf{Dataset} & \textbf{Model} \\ \midrule
AI Scientist~\cite{lu2024ai} & \gmark & \rmark & \gmark & \gmark & \rmark & \rmark & \gmark & \rmark & \rmark & \rmark & \rmark \\
AI Scientist v2~\cite{yamada2025ai} & \gmark & \rmark & \gmark & \gmark & \rmark & \rmark & \gmark & \rmark & \rmark & \rmark & \rmark \\
AI Researcher~\cite{tang2025ai} & \gmark & \rmark & \gmark & \ymark & \rmark & \rmark & \gmark & \gmark & \rmark & \rmark & \rmark \\
Agent Laboratory~\cite{schmidgall2025agent} & \gmark & \ymark & \gmark & \ymark & \rmark & \rmark & \gmark & \gmark & \rmark & \rmark & \rmark \\
TinyScientist~\cite{yu2025tinyscientist} & \gmark & \rmark & \gmark & \gmark & \rmark & \gmark & \gmark & \gmark & \gmark & \rmark & \rmark \\
DeepAnalyze~\cite{zhang2025deepanalyze} & \rmark & \gmark & \gmark & \ymark & \rmark & \rmark & \gmark & \gmark & \rmark & \gmark & \gmark \\
InternAgent~\cite{team2025internagent} & \gmark & \rmark & \gmark & \rmark & \rmark & \rmark & \gmark & \rmark & \rmark & \rmark & \rmark \\
InternAgent-1.5~\cite{feng2026internagent} & \gmark & \rmark & \gmark & \rmark & \rmark & \rmark & \gmark & \rmark & \rmark & \rmark & \rmark \\
\midrule
\sciagent (Ours) & \gmark & \gmark & \gmark & \gmark & \gmark & \gmark & \gmark & \gmark & \gmark & \gmark & \gmark \\
\bottomrule
\end{tabularx}
\label{tab:comparison}
\end{table*}

To address this fundamental disconnect, we present \textbf{\sciagent}, a \textbf{Sci}entific \textbf{D}ata-centric \textbf{E}nd-to-end \textbf{R}esearch system designed to flexibly automate the scientific research lifecycle. 
Distributed as a modular Python package, \sciagent integrates ideation, data analysis, and experimentation into a cohesive workflow.
The system consists of four sub-agents: the \emph{ideation} agent creates hypotheses and plans; the \emph{data analysis} agent cleans data and generates reports; the \emph{experimentation} agent writes and executes code; and the \emph{critic} agent evaluates all outputs to suggest improvements.
We equip each agent with a multimodal skill system, enabling it to adapt to diverse domains when solving interdisciplinary tasks.

\sciagent employs a data-centric approach to scientific discovery, independently parsing and analyzing raw experimental data across diverse domains to ensure consistency across data structures, quality, semantics, and experimentation.
While general coding assistants often struggle to bridge the gap between abstract reasoning and executable code in multidisciplinary fields like physics, biology, and remote sensing, \sciagent excels at solving complex, research-level problems by linking visual and structural data characteristics directly to code synthesis.
Furthermore, to reduce the community's reliance on closed-source APIs for these complex workflows, our framework empowers open-source models to achieve high scientific autonomy.
Qualitative feedbacks from experts and case studies also demonstrate that \sciagent effectively handles complex research tasks and enhances research capabilities. 

The primary contributions of this work are as follows:
\begin{itemize}
    \item We introduce \sciagent, a modular system of multimodal autonomous agents that automates the entire research lifecycle. \sciagent proposes a data-centric approach that grounds experiment-code generation in autonomous experimental analysis, enabling superior performance on interdisciplinary research problems.
    \item We release the \opensci-SFT-8K trajectory dataset for general research and the \opensci-27B model to advance scientific discovery with \sciagent.
    \item Empirical results demonstrate that our work outperforms existing baselines, proving its efficacy in complex scientific reasoning and coding tasks.
\end{itemize}

\section{Related Work}

\subsection{Autonomous Scientific Research Agents}
Early end-to-end automated pipelines, such as the AI Scientist series~\cite{lu2024ai,yamada2025ai} demonstrate that LLM-based agentic pipelines can generate hypotheses and run experiments with limited human intervention. Despite this progress, existing approaches do not provide support for the full scientific pipeline (Table~\ref{tab:comparison}). In particular, AI researcher~\cite{tang2025ai}, Agent Laboratory~\cite{schmidgall2025agent}, TinyScientist~\cite{yu2025tinyscientist}, and InternAgent~\cite{team2025internagent,feng2026internagent} focus primarily on idea generation and experiment execution for research. 
AI researcher introduces a fully autonomous pipeline for an open-ended research task. 
TinyScientist targets usability through an interactive and controllable framework; therefore is not fully autonomous as it requires continuous user involvement. 
Agent Laboratory and InternAgent integrate literature review, experimentation, and feedback refinement, yet lack an autonomous data analysis phase before experimental design.
Table~\ref{tab:comparison} compares \sciagent with prior research-agent
frameworks across functionality, capability, deployment, and open-source contribution.




\subsection{Evaluation Benchmarks for Research Agents}



To comprehensively evaluate our automated scientific workflow, we leverage several benchmarks across the research pipeline. For ideation, AI-Idea-Bench~\cite{qiu2025ai} measures an agent's ability to propose novel, feasible directions against 3,495 top-tier papers. For data analysis, DiscoveryBench~\cite{majumder2024discoverybench} tests multi-step statistical reasoning across 264 tasks. For experimentation and coding, MLE-Bench~\cite{chan2025mle} evaluates ML engineering on 75 Kaggle competitions, while SciCode~\cite{tian2024scicode} assesses scientific coding across 80 multi-step problems. To evaluate end-to-end autonomy, AIRS-Bench~\cite{lupidi2026airsbench} covers the entire research lifecycle across 20 complex tasks without baseline code. Finally, we use AstroVisBench~\cite{joseph2025astrovisbench} to assess multimodal reasoning and tool use, testing long-tail API integration and iterative visualization refinement across 864 tasks.

\begin{figure*}[t]
\centering
\includegraphics[width=\linewidth]{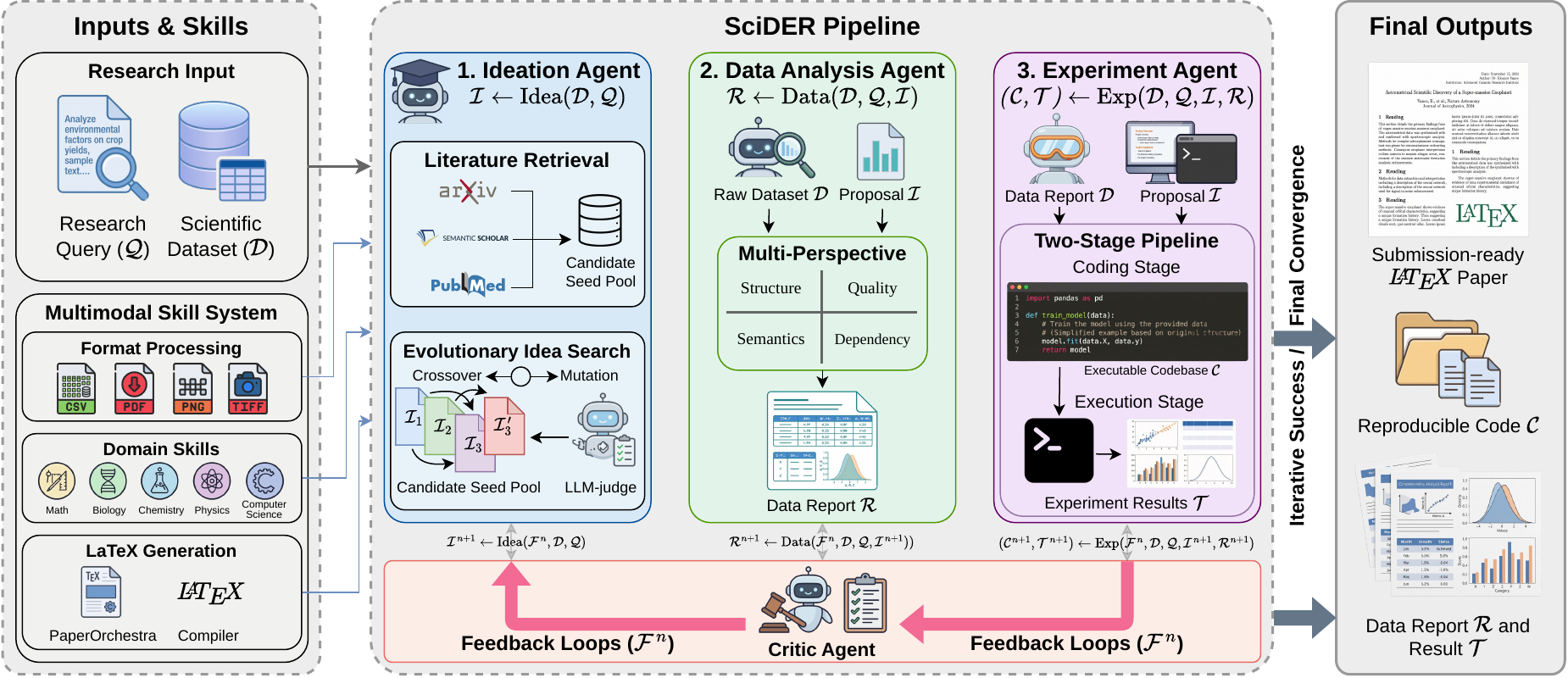}
\caption{
The system architecture of \sciagent. The pipeline automates the research lifecycle through four multimodal specialized agents: Ideation, Data Analysis, Experimentation, and Critic.
Inputs and skills are used to produce the final papers, code, and experimental results.
}
\label{fig:overview}
\end{figure*}

\section{System Architecture and Workflow}

\sciagent is designed as a data-centric end-to-end system that flexibly automates the scientific research lifecycle.
The system integrates a research framework comprising ideation, data analysis, experimentation, and iterative improvement. It supports flexible inputs such as text, raw data, code, and prior papers and codebases.
\sciagent also offers a lightweight web interface where researchers can upload their data and research topics, allowing the system to automatically create a closed-loop research cycle to propose and verify new ideas. 
Fig.~\ref{fig:overview} outlines the system design and provides an overview of modules and workflows.

The automated research task can be formalized as follows: given datasets $\mathcal{D}$ and a research query $\mathcal{Q}$, the system generates ideas $\mathcal{I}$ (i.e., hypothesis and experiment outlines), a data analysis report $\mathcal{R}$, an experiment codebase $\mathcal{C}$, and the final experimental results $\mathcal{T}$.

\subsection{LLM-based Agents}

\paragraph{\textbf{Ideation}: $\mathcal{I} \leftarrow \mathrm{Idea}(\mathcal{D}, \mathcal{Q})$} SciDER's ideation agent runs literature retrieval over arXiv~\cite{arxiv}, Semantic Scholar~\cite{semanticscholar}, and PubMed~\cite{pubmed}, then turns a seed pool of candidate ideas into a refined proposal using \textbf{Evolutionary Idea Search} (EIS). 
Since absolute LLM-judge~\cite{zheng2023judging} scores are noisy and poorly calibrated, selecting ideas from a small pool is effectively random.
EIS addresses this by prompting judges to rank K-way batches based on novelty, feasibility, impact, and specificity. These ranks are converted into a weighted composite score, and the population evolves using dimension-targeted improvement and rank-proportional combination operators within a fixed LLM-call budget. Top candidates are retained across generations. 
Each proposal includes a research hypothesis, an experimental outline, and a comparison to prior work.

EIS takes a pool of $n$ seed ideas $S = \{s_1, \dots, s_n\}$ produced
by literature-conditioned LLM ideation and returns a refined idea
$\mathcal{I}^\star$ along with provenance for every node in the search
tree. Two design choices are central.

The first design is \emph{K-way batch ranking}. For each evaluation dimension $d \in \mathcal{D} = \{\text{novelty}, \text{feasibility}, \text{impact}, \text{specificity}\}$, we prompt an LLM-judge with a dimension-specific question and ask it for a no-ties permutation $\pi_d$ over the current population indices. Ranks are converted to normalised scores $\mathrm{score}_d(x) = (n+1 - \mathrm{rank}_d(x))/n$ and combined into $S(x) = \sum_d w_d \cdot \mathrm{score}_d(x)$, with weights $w = (0.30, 0.25, 0.25, 0.20)$. The four rankings are issued in parallel. Asking for an ordering rather than absolute scores cancels the run-to-run calibration drift that makes absolute LLM scores unreliable for selection.

The second design is \emph{Evolutionary operators}. During each idea generation:
\begin{enumerate*}
\item Sort by composite, retain the top $k$ survivors, and pin the best seed so it cannot be displaced by re-ranking against a smaller batch.
\item Generate $n_{\mathrm{imp}}$ \emph{improve} children: each child takes a parent and its weakest dimension and rewrites the idea under a dimension-specific instruction.
\item Generate $n_{\mathrm{com}}$ \emph{combine} children: parent pairs are sampled by rank-proportional weighting and synthesised into a new idea.
\item Re-rank the merged batch on all four dimensions.
\end{enumerate*}
The loop terminates at an LLM-call budget or iteration cap, then runs one final calibrated pass so every final candidate is scored in the same $n$-way context. Algorithm~\ref{alg:eis} summarises the procedure. Default configuration: $n{=}8$, $k{=}4$, $T{=}3$, budget $60$ calls, $n_{\mathrm{imp}}/n_{\mathrm{com}} = 3/1$.

\begin{algorithm}[tbp]
\small
\caption{Evolutionary Idea Search (EIS)}
\label{alg:eis}
\begin{algorithmic}[1]
\Require seeds $S$, dimensions $\mathcal{D}$, weights $w$,
budget $B$, iterations $T$, survivors $k$
\State $P \gets S$;\ $\sigma \gets \textsc{BatchRank}(P, \mathcal{D}, w)$
\State $s^\star \gets \arg\max_{x \in S} \sigma(x)$
\Comment{pinned seed}
\For{$t = 1, \dots, T$}
  \State $P' \gets \textsc{Top}_k(P, \sigma) \cup \{s^\star\}$
  \State $P_{\mathrm{imp}} \gets \textsc{Improve}(P', \mathcal{D}, \sigma)$
  \State $P_{\mathrm{com}} \gets \textsc{Combine}(P', \sigma)$
  \State $P \gets P' \cup P_{\mathrm{imp}} \cup P_{\mathrm{com}}$
  \State $\sigma \gets \textsc{BatchRank}(P, \mathcal{D}, w)$
  \If{\textsc{Calls} $> B$} \textbf{break} \EndIf
\EndFor
\State \Return $\arg\max_{x \in P} \sigma(x)$, $P$, $\sigma$
\end{algorithmic}
\end{algorithm}

\paragraph{\textbf{Data Analysis}: $\mathcal{R} \gets \mathrm{Data}(\mathcal{D},\mathcal{Q},\mathcal{I})$}

This stage forms the core of \sciagent's data-centric paradigm, transforming raw scientific data into structured knowledge.
We model the input as a labelled file tree $\mathcal{T}_\mathcal{D} = (V, E, \tau)$, where $V$ enumerates files and directories under the workspace root, $E$ encodes parent--child containment, and the labelling map $\tau: V_{\mathrm{leaf}} \to \Sigma$ assigns each leaf a format from a registry $\Sigma$ (Parquet, CSV, TIFF, etc.). 
The agent visits $\mathcal{T}_\mathcal{D}$ in depth-first order, invoking a format-specific probe for each leaf $v$:
\begin{equation*}
\phi_{\tau(v)} : \mathrm{bytes}(v) \;\longrightarrow\; \mathrm{Schema}(v) \times \mathrm{Stats}(v),
\end{equation*}
which returns a typed schema and a fixed-shape statistical fingerprint (cardinality, missing rate, value range, distribution moments). Probes are pure functions of file content and are therefore cacheable across reruns.

The data report is the four-tuple
\begin{equation*}
\mathcal{R} \;=\; \big(\mathcal{R}_{\mathrm{str}},\ \mathcal{R}_{\mathrm{qual}},\ \mathcal{R}_{\mathrm{sem}},\ \mathcal{R}_{\mathrm{dep}}\big),
\end{equation*}
each component aggregating the per-leaf probes through a different lens:
\begin{enumerate*}
\item \emph{Structure} collects formats, dtypes, and schemas, $\mathcal{R}_{\mathrm{str}} = \{(v, \tau(v), \phi_{\tau(v)}(v))\}_{v \in V_{\mathrm{leaf}}}$;
\item \emph{Quality} summarises issues per field $f$ via the missing rate $\mu_f = |\{x \in f : x = \bot\}| / |f|$, outlier mass under a robust $z$-score threshold, and constraint violations;
\item \emph{Semantics} binds each field to a query-conditioned role $\psi(f \mid \mathcal{Q}, \mathcal{I}) \in \{\text{target}, \text{time-index}, \text{covariate}, \text{identifier}, \dots\}$, so downstream code addresses fields by role rather than column name;
\item \emph{Dependency} extracts a graph $G_{\mathrm{dep}} = (F, E_{\mathrm{dep}})$ over fields $F$, with edges for matched primary/foreign keys, shared timestamps, and value-set overlap above a tunable threshold.
\end{enumerate*}
While the pipeline may automatically preprocess and clean the data (yielding $\mathcal{D}'$), we denote the input space as $\mathcal{D}$ for simplicity.
The structured report $\mathcal{R}$ serves as the foundation for the \emph{experiment agent}, guiding data loading, preprocessing, and field selection during experiments.


\paragraph{\textbf{Experimentation}:  $(\mathcal{C}, \mathcal{T}) \gets \mathrm{Exp}(\mathcal{D},\mathcal{Q},\mathcal{I},\mathcal{R})$}
The experimentation stage interleaves code generation and execution. Guided by the proposal $\mathcal{I}$ and data report $\mathcal{R}$, the agent generates an executable codebase $\mathcal{C}$ tailored to the domain's data structures and dependencies. This codebase is run in an isolated workspace to produce results $\mathcal{T}$. We model this workspace as a state $\mathcal{W} = (\mathcal{C}, \mathcal{D}, \mathcal{L})$, comprising the codebase, the preprocessed data $\mathcal{D}$, and an execution log $\mathcal{L}$ capturing stdout, stderr, and test artifacts.

\emph{Coding phase:} $\mathcal{C} \gets \mathrm{Coding}(\mathcal{D},\mathcal{Q},\mathcal{I},\mathcal{R})$.
The coding agent iteratively applies patches $\Delta_k$ to the workspace, validating them with a static guard $g(\mathcal{C}_k) \in \{0, 1\}$ (syntax, type, and lint checks). If validation fails, the agent reflects on the error trace to generate the next patch  $\Delta_{k+1}$~\cite{chen2024selfdebug}:
\begin{equation*}
\mathcal{C}_{k+1} =
\begin{cases}
\mathcal{C}_k & \text{if } g(\mathcal{C}_k) = 1, \\
\mathcal{C}_k \oplus \Delta_{k+1} & \text{otherwise},
\end{cases}
\quad k = 0, \dots, K_{\max}{-}1,
\end{equation*}
where $\oplus$ denotes patch application and $K_{\max}$ is the budget. The phase commits once $g(\mathcal{C}_k) = 1$ or the budget is exhausted.

\emph{Execution phase:} $\mathcal{T} \gets \mathrm{Exec}(\mathcal{D}',\mathcal{C})$.
After passing the static guard, the execution agent launches $\mathcal{C}$ and periodically samples a runtime signal $\eta_t = (\ell_t, \mathrm{loss}_t, \mathrm{prog}_t)$ at time $t$, where $\ell_t$ is the latest log tail, $\mathrm{loss}_t$ is the most recent training loss (if any), and $\mathrm{prog}_t \in [0, 1]$ is a coarse progress estimate. The agent typically runs to completion, but if a stopping predicate fires, it terminates and emits a structured failure $\bot$ to the coding agent. This structured failure feedback ($\eta_t$) is sent back to the coding phase to enable a targeted code revision cycle.

\paragraph{\textbf{Critics \& Feedback}: $\mathcal{F} \gets \mathrm{Critic}(\cdots)$}

After each analysis and experimentation phase, a \textit{critic agent} evaluates intermediate outputs to drive iterative self-refinement, assessing the workflow for \emph{accuracy} (eliminating misinformation and hallucinations), \emph{completeness} (resolving technical and information gaps), and \emph{neutrality} (filtering bias).

Formally, let $x^n = (\mathcal{I}^n, \mathcal{R}^n, \mathcal{C}^n, \mathcal{T}^n)$ denote the artifacts at iteration $n$. The critic emits
\begin{equation*}
\mathcal{F}^n = \mathrm{Critic}(x^n) = \Big(v^n,\ \{(d,\, \sigma_d^n,\, \delta_d^n)\}_{d \in \mathcal{D}_{\mathrm{crit}}}\Big),
\end{equation*}
where $\mathcal{D}_{\mathrm{crit}}$ comprises the three axes above, $\sigma_d^n \in [0,1]$ is the per-axis confidence, $\delta_d^n$ is a natural-language patch instruction, and $v^n \in \{\textsc{pass}, \textsc{revise}\}$ is the gating verdict. Instead of merely flagging errors, the critic provides $\delta_d^n$ as concrete revision targets, enabling downstream agents to revise locally without re-running unaffected stages.

The verdict is finalized by an \emph{approval} layer, $v^n = \mathrm{Approve}(x^n, \mathcal{F}^n; \mathcal{D}_{\text{ws}})$, where $\mathcal{D}_{\text{ws}}$ is the workspace state. Interactively, this layer presents $\mathcal{F}^n$ to a domain expert via a web interface; in headless runs, an \textit{approval subagent} verifies the critic's claims against $\mathcal{D}_{\text{ws}}$, mitigating the known instability of single-pass LLM-as-a-judge scoring~\cite{zheng2023judging}.
Given the committed verdict, the framework generates the next revision:
{
\begin{align*}
\mathcal{I}^{n+1} &\gets \mathrm{Idea}(\mathcal{F}^n, \mathcal{D},\mathcal{Q}) \\
\mathcal{R}^{n+1} &\gets \mathrm{Data}(\mathcal{F}^n,\mathcal{D},\mathcal{Q},\mathcal{I}^{n+1}) \\
(\mathcal{C}^{n+1},&\mathcal{T}^{n+1}) \gets \mathrm{Exp}(\mathcal{F}^{n},\mathcal{D},\mathcal{Q},\mathcal{I}^{n+1},\mathcal{R}^{n+1}) \\
\mathcal{F}^{n+1} &\gets \mathrm{Critic}(\mathcal{F}^{n},\mathcal{I}^{n+1},\mathcal{R}^{n+1},\mathcal{C}^{n+1},\mathcal{T}^{n+1})
\end{align*}
}
To prevent pipeline stalls from pathological reject cycles, the loop terminates once $v^n = \textsc{pass}$ or the per-agent retry budget is exhausted (default $N_{\max}{=}2$).

\subsection{Multimodal Skill System}

To leverage community-driven AI, we implemented a flexible skill system that dynamically loads capabilities into the context. Inspired by tool-augmented LLM agents~\cite{qin2024toolllm}, it operates at the granularity of \emph{procedural recipes} rather than single tool calls.
Formally, a skill is a tuple
\begin{equation*}
s \;=\; \langle \mathrm{name},\ \mathrm{desc},\ A_s,\ P_s,\ B_s \rangle,
\end{equation*}
where $A_s \subseteq \mathcal{A}$ restricts $s$ to a subset of agents $\mathcal{A}$, $P_s \subseteq A_s$ specifies which agents preload $s$ in their system prompt, and $B_s$ is the on-demand skill body. The registry $\mathcal{S}$ is auto-scanned from designated folders, presenting agent $a$ with a role-restricted view.
\begin{equation*}
\mathcal{S}_a \;=\; \{ s \in \mathcal{S} : a \in A_s \}.
\end{equation*}

At inference, agent $a$ initially incurs a cost of only $\sum_{s \in \mathcal{S}_a} |\mathrm{desc}(s)|$ tokens by loading short skill descriptors instead of their full bodies. A skill $s$ is materialized only when triggered by keyword matching against the user query $\mathcal{Q}$ or explicitly invoked via $\mathrm{Skill}(s) \to B_s$. Letting $\mathcal{K}_t \subseteq \mathcal{S}_a$ denote the skills invoked up to turn $t$, the prompt footprint is
\begin{equation*}
\mathrm{Tok}_t(a) \;=\; \underbrace{\sum_{s \in \mathcal{S}_a} |\mathrm{desc}(s)|}_{\text{catalogue (linear in $|\mathcal{S}_a|$)}}
\;+\; \underbrace{\sum_{s \in \mathcal{K}_t} |B_s|}_{\text{materialised bodies}},
\end{equation*}
decoupling per-call costs from catalogue size ensures that adding new skills costs negligible tokens until activated. 

This design integrates advanced research capabilities without altering the core architecture:
\begin{enumerate*}[label=(\arabic*)]
    \item Multimodal skill support for processing diverse file formats, including multimedia documents and raster images.
    \item Domain-specific skills spanning mathematics, statistics, biology, ecology, chemistry, materials, physics, social sciences, and computer science.
    \item Generation of submission-ready \LaTeX~papers via PaperOrchestra~\cite{song2026paperorchestra}.
\end{enumerate*}

\subsection{\opensci}

To democratize autonomous scientific discovery, we curated \textsc{\opensci-SFT-8K}, a dataset comprising more than 8K high-quality execution trajectories as summarized in Table~\ref{tab:sft-by-agent}. 
We curate 2,678 trajectories from a comprehensive suite of scientific benchmarks (e.g., DataSciBench~~\cite{zhang2025datascibench}, DS-1000~\cite{lai2023ds}, DS-Bench~\cite{jing2024dsbench}, and ScienceAgentBench~\cite{chen2025scienceagentbench}) using Qwen-3.6-27B.
Additionally, the dataset includes 5,854 successful benchmark trajectories for future use.
We confirmed that no evaluation trajectories from Section~\ref{sec:evaluation} were used in training by separating them into distinct subsets.
To maximize learning density, our preprocessing pipeline merged consecutive same-role messages, capped tool outputs at 512 tokens, and split trajectories at user turns into 16,384-token segments.

We fine-tuned the Qwen-3.6-27B backbone~\cite{qwen2026qwen36} on this dataset to create \opensci-27B, our open-weights model. Training ran for two epochs using LoRA~\cite{hu2022lora} ($r=32, \alpha=64$) on all linear projections, optimized via AdamW (peak learning rate $1 \times 10^{-4}$, cosine schedule), and DeepSpeed ZeRO-2~\cite{rajbhandari2020zero} in BF16 using 2$\times$ H200 GPUs. The merged LoRA adapter is deployed via vLLM with YaRN~\cite{peng2024yarn} scaling to support up to a 512K context window. Both the model and dataset are released under the Apache-2.0 license.

\begin{table}[t]
\centering
\caption{Statistics of \textsc{\opensci-SFT-8K} trajectories by agent role. 
}
\label{tab:sft-by-agent}
\newcolumntype{C}{>{\centering\arraybackslash}X}
\begin{tabularx}{0.8\linewidth}{@{}C ccc@{\quad}}
\toprule
\textbf{Agent role}            & \textbf{\#Traj.} & \textbf{Tokens}   & \textbf{Tok./traj.} \\
\midrule
\multicolumn{4}{c}{\textit{Main agents}} \\
\quad \textsc{experiment}      &    947  & 235.0\,M & 248.2\,k \\
\quad \textsc{data}            &    555  & 118.8\,M & 214.0\,k \\
\quad \textsc{ideation}        &    430  &  15.0\,M &  34.9\,k \\
\midrule
\multicolumn{4}{c}{\textit{Subagents}} \\
\quad \textsc{coding}          & 2{,}610 & 122.6\,M &  47.0\,k \\
\quad \textsc{critic}          & 2{,}119 &  91.5\,M &  43.2\,k \\
\quad \textsc{approval}        & 1{,}863 & 147.5\,M &  79.2\,k \\
\quad \textsc{paper\_search}   &      8  &   0.1\,M &   6.8\,k \\
\midrule
\textbf{Total}                 & 8{,}532 & 730.5\,M &  85.6\,k \\
\bottomrule
\end{tabularx}
\end{table}

\section{Evaluation}
\label{sec:evaluation}


\subsection{Evaluation Setup}
Our evaluation uses existing benchmarks, supplemented by human feedback and case studies. 
The evaluation spans ideation, data analysis, and experimentation. We also assess \sciagent's multimodal performance and ablate its ideation and data analysis modules.
Results for other methods and models are sourced directly from their benchmarks and leaderboards, where available.
We follow these benchmark implementations to ensure rigorous benchmark results.

\subsection{Idea Generation}

To rigorously evaluate the ideation phase of \sciagent, we use the AI-Idea-Bench 2025~\cite{qiu2025ai} framework, which provides a comprehensive and quantitative method for assessing AI-generated research ideas. 
This benchmark evaluates generated ideas using 3,495 top-tier AI papers, assessing ground-truth alignment and objective, and reference-based judgment.
Specifically, it evaluates ideas across three key dimensions: (1) \textbf{Quality} (Idea-to-Idea Matching), which scores the conceptual similarity of the motivation and experimental design to target papers on a 0--5 scale; (2) \textbf{Novelty}, which quantifies originality by measuring the distance to existing literature, weighted by citation impact; and (3) \textbf{Feasibility}, which assesses methodological grounding based on the citation influence of reference papers.
Baseline results of SCIPIP~\cite{wang2024scipip}, VIRSC~\cite{su2025many}, AI-Researcher, and AI-Scientist v1 \& v2, and InternAgent-1.5~\cite{feng2026internagent} are also reported using Gemini-2.5-Pro~\cite{comanici2025gemini} as well as \sciagent. 
Table \ref{tab:aiideabench} presents our results. Unless otherwise noted, \textbf{bold} and \underline{underlined} text indicate the first and second best results, respectively, throughout this document.

Table~\ref{tab:aiideabench} shows that \opensci significantly outperforms InternAgent-1.5 in novelty, scoring 63.22 (+12.98\%) in motivation and 56.11 (+6.41\%) in experiment. 
The EIS module also boosts \opensci’s Feasibility Score to 36.2, marking a 17.15\% gain over InternAgent-1.5. Furthermore, \opensci surpasses both the standard \sciagent and AI-Scientist-v2 across all metrics, confirming the effectiveness of our core mechanisms.
These results show that EIS generates innovative and executable research proposals.

\begin{table}[tbp]
\centering
\newcolumntype{C}{>{\centering\arraybackslash}X}
\caption{
Results of \texttt{AI-Idea-Bench}. 
}
\begin{tabularx}{\linewidth}{@{} c CC CC C @{}}
\toprule
\multirow{2}{*}{\textbf{Method}} & \multicolumn{2}{c}{\textbf{Quality}} & \multicolumn{2}{c}{\textbf{Novelty}} & \multirow{2}{*}{\textbf{Feas.}} \\
\cmidrule(lr){2-3} \cmidrule(lr){4-5}
 & Motiv. & Exp. & Motiv. & Exp. &  \\ 
\midrule
SCIPIP~\cite{wang2024scipip} & 2.44 & - & 25.06 & - & - \\
VIRSC~\cite{su2025many} & 2.94 & 2.12 & 24.87 & 24.65 & 13.3 \\
AI-Researcher~\cite{tang2025ai} & 3.56 & 3.02 & 34.92 & 34.69 & 18.3 \\
AI-Scientist~\cite{lu2024ai} & 3.63 & 3.28 & 39.03 & 36.08 & 15.7 \\
AI-Scientist-v2~\cite{yamada2025ai} & 4.24 & 3.71 & 44.77 & 42.31 & 22.5 \\
InternAgent-1.5~\cite{feng2026internagent} & \textbf{4.62} & \textbf{4.15} & 55.96 & 52.73 & 30.9 \\
\midrule
\sciagent & 4.32 & 3.83 & 47.53 & 46.52 & 26.1 \\
\opensci & \underline{4.45} & \underline{3.96} & \textbf{63.22} & \textbf{56.11} & \textbf{36.2} \\ 
\bottomrule
\end{tabularx}
\label{tab:aiideabench}
\end{table}

\subsection{Data Analysis}



To systematically evaluate the agentic data analysis capabilities of our framework, we employ DiscoveryBench~\cite{majumder2024discoverybench}, the first comprehensive benchmark that formalizes the multi-step process of data-driven discovery. It features 264 real-world tasks spanning six diverse domains, manually derived from published scientific workflows. Each task requires the agent to integrate raw datasets, metadata, and high-level discovery goals to conduct programmatic exploration and statistical reasoning. We compare our approach against state-of-the-art baselines, including DataVoyager~\cite{majumder2024data} and Asta-v0~\cite{allenai2026asta}.

Table~\ref{tab:discoverybench} shows that \sciagent achieves state-of-the-art performance in complex data-driven hypothesis testing. Powered by Claude-Sonnet-4~\cite{anthropic2025claude4systemcard}, it achieves a peak accuracy of \textbf{35.4\%}, outperforming the strongest baseline, Asta-v0, by 2.2 percentage points. Additionally, our smaller open-weights variant, \sciagent (\opensci-27B), remains highly competitive at \textbf{31.5\%} accuracy score.
Notably, it substantially outperforms the proprietary-model-backed DataVoyager (25.7\%) by a relative margin of 22.5\% and closely approaches Asta-v0. This highlights that our framework effectively empowers open-weight models to bridge the performance gap with massive proprietary LLMs in rigorous scientific data analysis tasks.

Detailed per-task performance breakdowns are provided in Fig.~\ref{fig:discoverybench}, further demonstrating \sciagent's consistent superiority across diverse scientific sub-domains.

\begin{table}[tbp]
\caption{
Results of \texttt{DiscoveryBench}.
}
\newcolumntype{C}{>{\centering\arraybackslash}X}
\begin{tabularx}{\linewidth}{@{}CcC@{}}
\toprule
\textbf{Method} & \textbf{Model} & \textbf{Accuracy} \\
\midrule
DataVoyager~\cite{majumder2024data} & Claude-Sonnet-4 & 0.257 \\
Asta-v0~\cite{allenai2026asta} & Claude-Sonnet-4 & 0.332 \\
\midrule
\sciagent & Gemini-3-Flash & 0.274 \\
\sciagent & Claude-Sonnet-4 & \textbf{0.354} \\
\sciagent & \opensci-27B & 0.315 \\
\bottomrule
\end{tabularx}
\label{tab:discoverybench}
\end{table}

\subsection{Experimentation}



To evaluate \sciagent's autonomous execution and code synthesis capabilities, we use two benchmarks: MLE-Bench~\cite{chan2025mle} and SciCode~\cite{tian2024scicode}. MLE-Bench measures machine learning proficiency on Kaggle-style challenges via medal-level achievements (\%Any and \%Gold). SciCode benchmark is a scientist-curated scientific benchmark comprising 80 core problems (decomposed into 338 sub-problems) across 16 subfields, testing knowledge recall, scientific reasoning, and complex code synthesis.

Table~\ref{tab:mlebench} demonstrates that \sciagent establishes a new state-of-the-art on the MLE-Bench \texttt{Lite} split. Powered by Gemini-3-Pro, it achieves \textbf{63.64\%} for any medal and \textbf{40.90\%} for gold medals, outperforming the strongest baseline, AIRA~\cite{toledo2025aira}, by absolute margins of 6.75\% and 3.74\%, respectively. Remarkably, our 27B open-weights variant, \sciagent (\opensci-27B), achieves a highly competitive 54.54\% overall medal rate. It also outperforms other agentic systems, such as ML-Master (48.50\%)~\cite{liu2025mlmaster} and AIDE (16.90\%)~\cite{jiang2025aide}.

\begin{figure}[t]
\centering
\includegraphics[width=\linewidth]{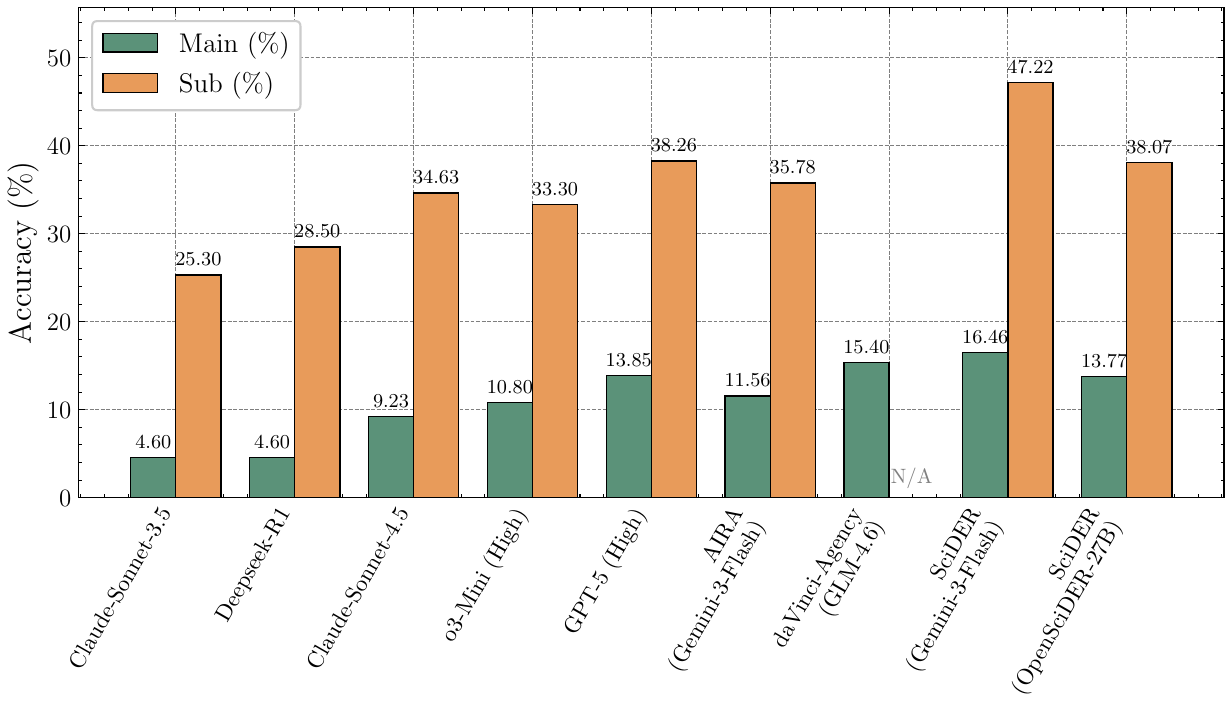}
\caption{
Results of \texttt{SciCode}. Solve rates for main and subproblems are reported, with higher rates indicating greater domain-specific capability.
}
\label{fig:scicode}
\end{figure}

Similar breakthroughs emerge in general scientific coding. As shown in Fig.~\ref{fig:scicode}, \sciagent (Gemini-3-Flash~\cite{gdm2025gemini3}) significantly outperforms AIRA by an absolute margin of \textbf{11.44\%} on sub-problem success (a striking 31.97\% relative improvement) and surpasses daVinci on main-problem success (16.46\% vs. 15.40\%). Notably, the lightweight \opensci-27B model again outperformed the proprietary AIRA (Gemini-3-Flash), achieving a 38.07\% success rate against AIRA’s 35.78\%. These consistent gains across two distinct benchmarks underscore the effectiveness of our research-centric design, proving that our iterative trial-critic-revision cycles ensure highly robust execution across diverse scientific domains.

\begin{table}[tbp]
\centering
\newcolumntype{C}{>{\centering\arraybackslash}X}
\caption{
Results of \texttt{MLE-Bench} on the \texttt{Lite} split. 
}
\begin{tabularx}{\linewidth}{@{}cc CC@{}}
\toprule
\multirow{2}{*}{\textbf{Method}} & \multirow{2}{*}{\textbf{Model}} & \multicolumn{2}{c}{\textbf{MLE-Bench (Lite)}} \\ \cmidrule{3-4} 
 & & \textbf{\%Any} $\uparrow$ & \textbf{\%Gold} $\uparrow$ \\ 
\midrule
AIDE~\cite{jiang2025aide} & o1-Preview & 16.90 & 9.40 \\
ML-Master~\cite{liu2025mlmaster} & Deepseek-R1 & 48.50 & 18.10 \\
AIRA~\cite{toledo2025aira} & o3-Preview & 47.73 & 28.64 \\
AIRA~\cite{toledo2025aira} & Gemini-3-Pro & 56.89 & 37.16 \\
\midrule
\sciagent & Gemini-3-Pro & \textbf{63.64} & \textbf{40.90} \\
\sciagent & \opensci-27B & 54.54 & 31.82 \\
\bottomrule
\end{tabularx}
\label{tab:mlebench}
\end{table}


\subsection{End-to-end Research}



We evaluated our framework's cross-domain, end-to-end research capabilities on AIRS-Bench~\cite{lupidi2026airsbench}, comparing it against standard ReAct~\cite{yao2022react} agents and the state-of-the-art AIRA framework. Unlike static coding tests, AIRS-Bench assesses the complete research lifecycle from ideation to refinement across $\sim$20 complex machine learning tasks that require agents to innovate without baseline code.

Table~\ref{tab:airsbench} demonstrates a significant leap in full-lifecycle autonomous research. Powered by Claude-Sonnet-4, \sciagent achieves a better score of \textbf{0.780}, outperforming the previous state-of-the-art (AIRA with GPT-5 at 0.550) by an absolute margin of 0.23 (a 41.8\% relative improvement). 
The results are collected within five different benchmark iterations.

Our lightweight open-weights variant, \opensci-27B, achieves a compelling score of \textbf{0.519}. This 27B model outperforms massive architectures like ReAct (GPT-OSS-120B, 0.402) and proprietary models like AIRA (o3, 0.424), while closely rivaling the flagship GPT-5. These results confirm that our end-to-end architecture raises the reasoning capabilities of open-weight models, enabling them to execute complex research workflows previously reserved for other systems. 
Fig.~\ref{fig:airsbench_pertask} shows per-task scores in details.

\begin{table}[tbp]
\centering
\newcolumntype{C}{>{\centering\arraybackslash}X}
\caption{
Results of \texttt{AIRS-Bench}.
}
\begin{tabularx}{\linewidth}{@{}CcC@{}}
\toprule
\textbf{Method} & \textbf{Model} & \textbf{Score} \\
\midrule
ReAct~\cite{yao2022react} & o3-Mini & 0.391 \\
ReAct~\cite{yao2022react} & GPT-OSS-120B & 0.402 \\
AIRA~\cite{toledo2025aira} & o3 & 0.424 \\
AIRA~\cite{toledo2025aira} & GPT-5 & 0.550 \\
\midrule
\sciagent & Claude-Sonnet-4 & \textbf{0.780} \\
\sciagent & \opensci-27B & 0.519 \\
\bottomrule
\end{tabularx}
\label{tab:airsbench}
\end{table}

\subsection{Visual Understanding}



To evaluate multimodal scientific reasoning and code generation for specialized tools, we benchmark our framework on AstroVisBench~\cite{joseph2025astrovisbench}. As the first domain-specific benchmark for end-to-end astronomical computing and visualization, it comprises 864 expert-curated tasks designed to rigorously test language models on long-tail API usage. In this setting, the agent must synthesize specialized code and iteratively refine complex plots utilizing its own visual feedback. 
Since existing scientific discovery agents lack robust multimodal support, we establish a standard multimodal ReAct agent as our baseline.

\begin{figure}[t]
\centering
\includegraphics[width=0.73\linewidth]{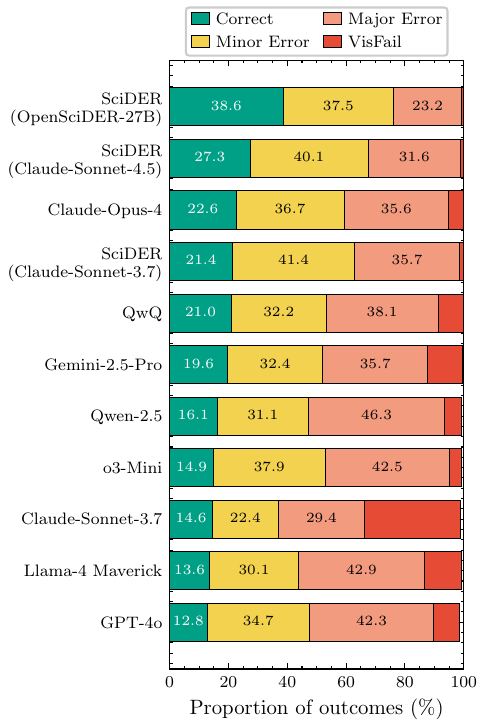}
\caption{
Results of \texttt{AstroVisBench}. Correct, Failure, Minor Error, and Major Error rates are reported.
}
\label{fig:astrovisbench}
\end{figure}

Fig.~\ref{fig:astrovisbench} illustrates a compelling dynamic in handling out-of-distribution scientific tasks. Notably, \opensci significantly outperforms both the standard multimodal ReAct agent and, remarkably, the closed-source-backed \sciagent framework. While frontier proprietary models often struggle with the highly specialized, long-tail API syntax required in astrophysics, our open-weights \opensci successfully leverages its targeted design to achieve a higher rate of correct visual outputs and substantially fewer major errors. This underscores our method's superior adaptability to highly specialized, visually grounded research workflows.


\subsection{Ablation Study}

\paragraph{Evolutionary Idea Search}
While the preceding evaluations demonstrate the overall superiority of our framework across diverse benchmarks, we conduct a targeted ablation study to isolate the specific impact of the EIS module.

To quantify the performance lift generated by EIS, we evaluate the system's output before and after the evolutionary process on AI-Idea-Bench. Because all dimensions are scored simultaneously, we log the initial best composite score ($\sigma_0^\star$) from the literature-conditioned seed pool as our natural baseline. We then measure the absolute lift ($\Delta = \sigma_T^\star - \sigma_0^\star$) achieved after $T$ evolutionary generations.

\begin{figure}[tbp]
\centering
\includegraphics[width=\linewidth]{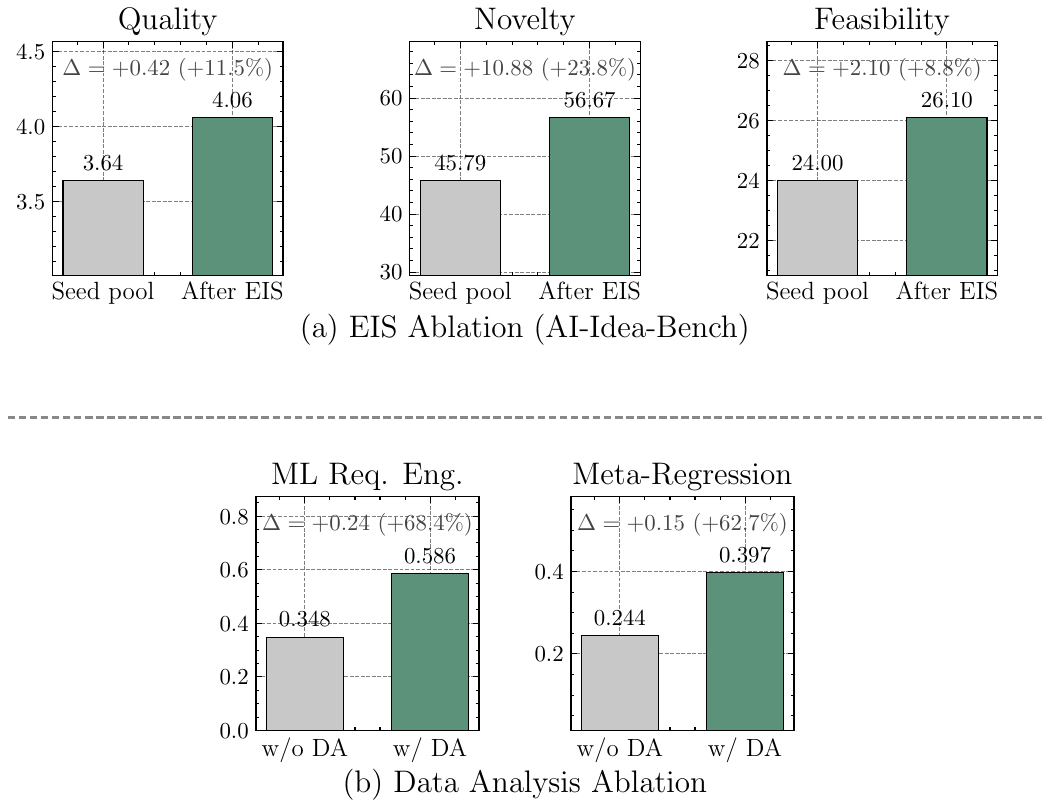}
\caption{
Ablation Study of EIS and Data Analysis:
(a) Impact of EIS on AI-Idea-Bench quality, novelty, and feasibility.
(b) Impact of Data Analysis on DiscoveryBench accuracy in different tasks.
}
\label{fig:ablation}
\end{figure}

Fig.~\ref{fig:ablation}(a) demonstrates that EIS is the critical driver of our framework's high performance. Relying solely on the initial seed pool yields mediocre results. However, applying the EIS operators drives a substantial \textbf{23.8\%} relative improvement in Novelty (from 45.79 to 56.67) and an \textbf{11.5\%} gain in Quality (from 3.64 to 4.06), while simultaneously enhancing Feasibility. These results indicate that initial LLM-generated ideas are often generic, and our iterative evolutionary search is essential for discovering truly innovative and rigorously grounded research proposals.


\paragraph{Data Analysis}
To verify the impact of our data-centric design on automated research, we conducted an ablation study on the data analysis module. Specifically, we evaluated the \sciagent system on the DiscoveryBench \texttt{ML Req. Eng.} and \texttt{Meta-Regression} task, with and without this module. As shown in Fig.~\ref{fig:ablation}(b), including the data analysis module improved the score by 0.238 and 0.153 on each task, confirming its effectiveness in downstream tasks.

\begin{figure*}[tb]
    \centering
    \newcommand{\myWidth}{0.185}

    \subfloat[Page 1\label{fig:scider-paper-p1}]{
        \includegraphics[
            page=1,
            width=\myWidth\textwidth,
            trim=8mm 8mm 8mm 8mm,
            clip
        ]{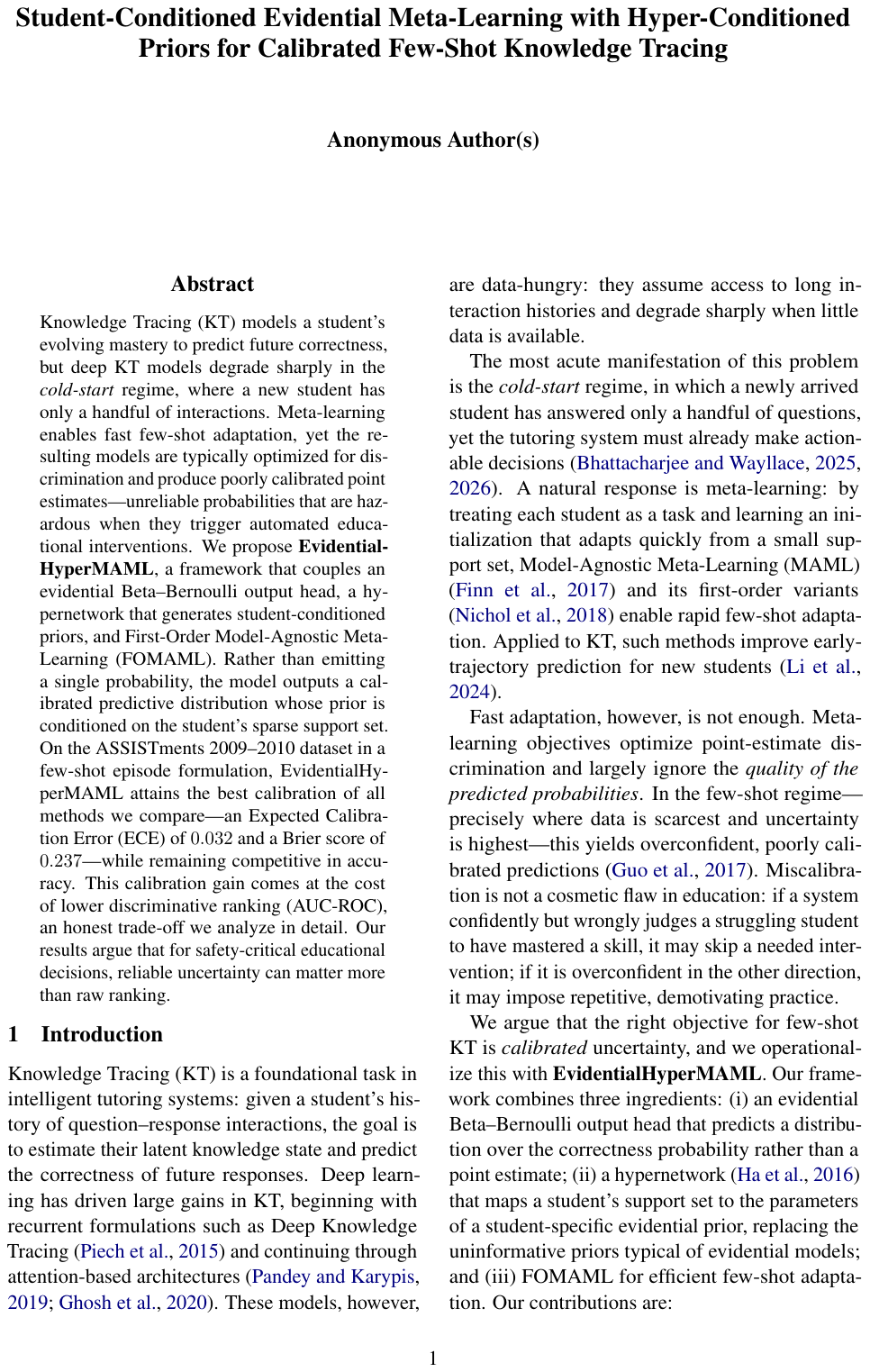}
    }
    \hfill
    \subfloat[Page 2\label{fig:scider-paper-p2}]{
        \includegraphics[
            page=2,
            width=\myWidth\textwidth,
            trim=8mm 8mm 8mm 8mm,
            clip
        ]{img/paper.pdf}
    }
    \hfill
    \subfloat[Page 4\label{fig:scider-paper-p4}]{
        \includegraphics[
            page=4,
            width=\myWidth\textwidth,
            trim=8mm 8mm 8mm 8mm,
            clip
        ]{img/paper.pdf}
    }
    \hfill
    \subfloat[Page 5\label{fig:scider-paper-p5}]{
        \includegraphics[
            page=5,
            width=\myWidth\textwidth,
            trim=8mm 8mm 8mm 8mm,
            clip
        ]{img/paper.pdf}
    }
    \hfill
    \subfloat[Page 6\label{fig:scider-paper-p6}]{
        \includegraphics[
            page=6,
            width=\myWidth\textwidth,
            trim=8mm 8mm 8mm 8mm,
            clip
        ]{img/paper.pdf}
    }

    \caption{
        Selected generated pages showcasing end-to-end scientific discovery evidence for \textit{``Student-Conditioned Evidential Meta-Learning with Hyper-Conditioned Priors for Calibrated Few-Shot Knowledge Tracing''}.
        The tiled pages show the generated paper, covering the method, experiments, quantitative results, calibration analysis, limitations, and ethics discussion.
    }
    \label{fig:scider-evidence}
\end{figure*}

\begin{figure*}[tb]
    \centering
    \subfloat[\label{fig:astrovis_2}]{
        \includegraphics[width=0.48\linewidth]{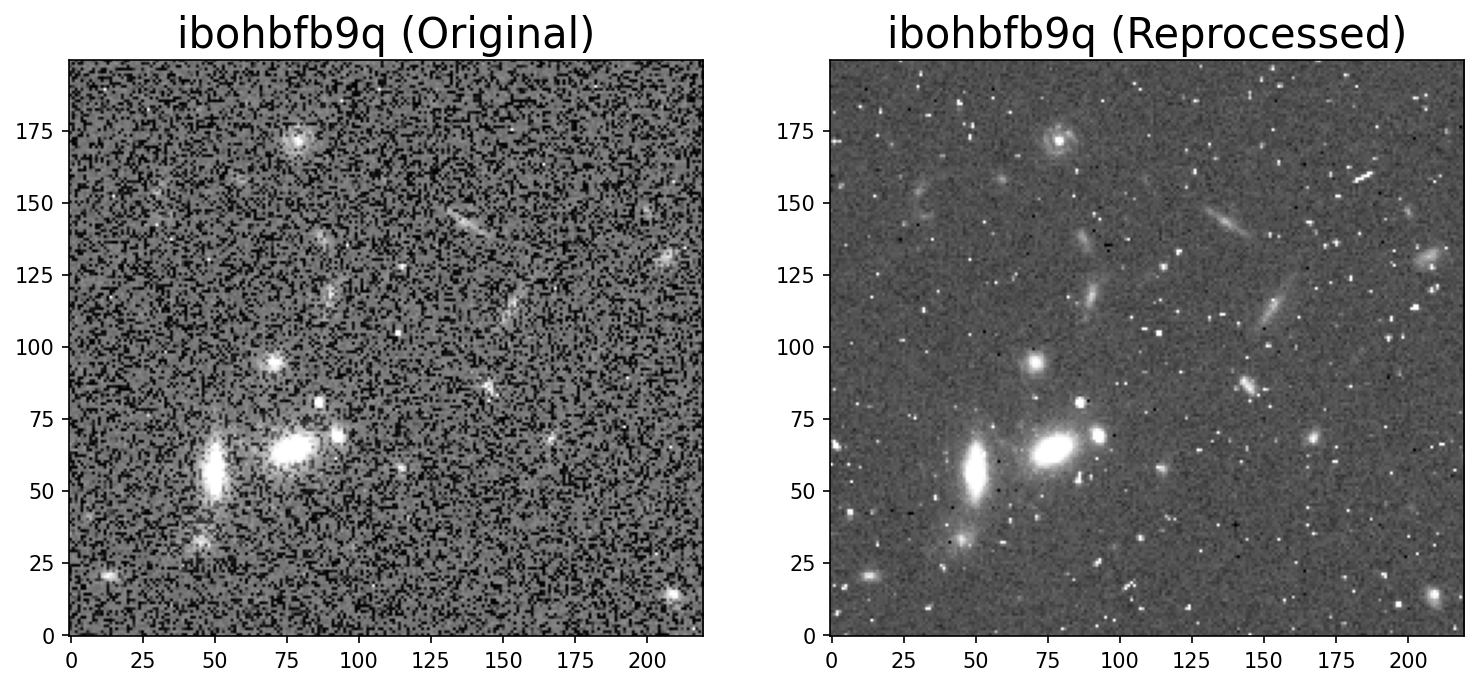}
    }
    \hfill
    \subfloat[\label{fig:astrovis_3}]{
        \includegraphics[width=0.48\linewidth]{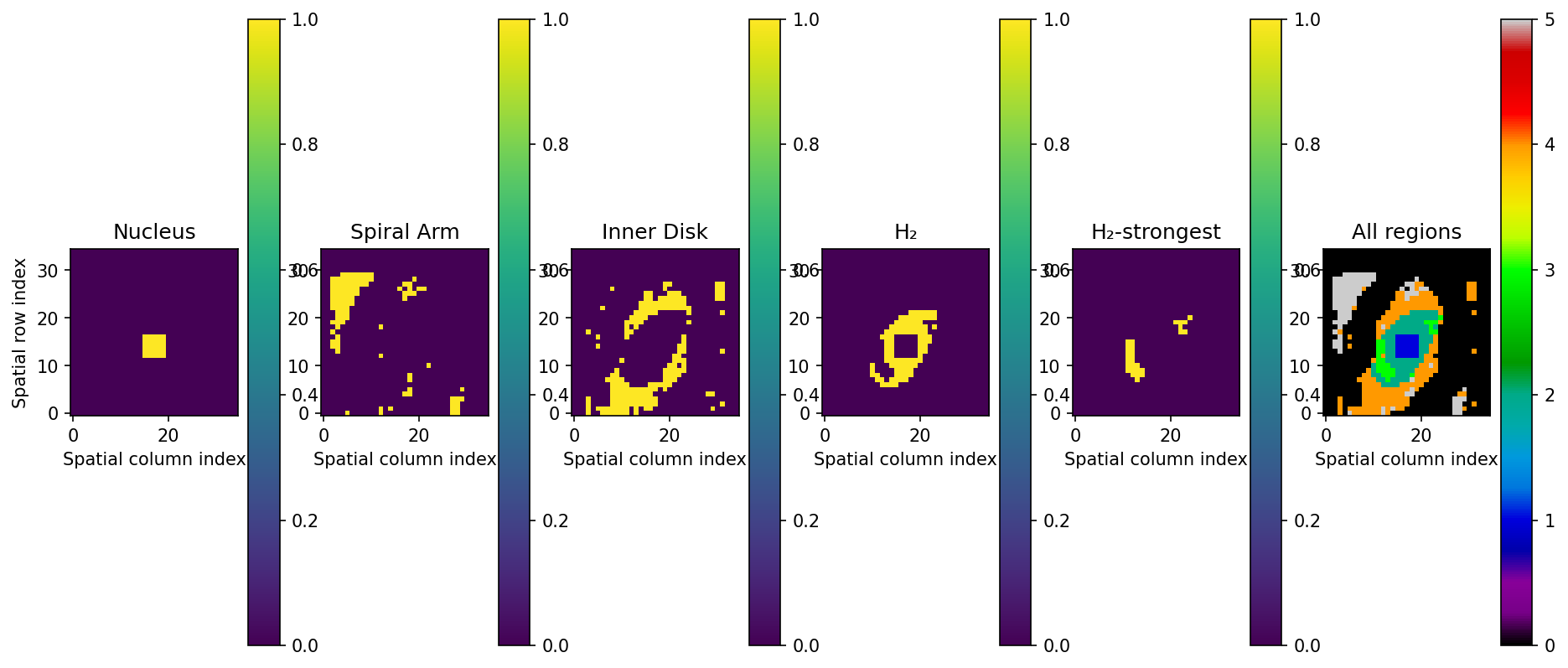}
    }
    
    \caption{Qualitative generated examples from \texttt{AstroVisBench}. Utilizing an iterative multimodal feedback loop, \sciagent successfully synthesizes and refines complex astronomical visualizations, accurately handling long-tail APIs and domain-specific formatting.}
    \label{fig:astrovis_examples}
\end{figure*}

\subsection{Qualitative Feedback}

To assess the practical utility of \sciagent, we collect qualitative feedbacks using a public questionnaire with 13 domain experts (professors, Ph.D. students, and industrial researchers). 
Participants are encouraged to try various workflows and review the outputs.
They also evaluate how effectively the system reduces the human researcher's workload from data parsing to experimental execution, using the 5-point ``helpfulness'' rubric detailed in Table~\ref{tab:human_rubric}.

\sciagent achieved a mean score of 4.846/5.000 with a standard deviation of 0.376. This consensus demonstrates that domain experts overwhelmingly classify the system near the ``Excellent'' tier, validating its efficacy as a highly autonomous and reliable scientific collaborator.

\begin{table}[htbp]
\centering
\caption{Rubric for qualitative feedback.}
\begin{tabularx}{\linewidth}{@{} l >{\hsize=1.2\hsize}X @{}}
\toprule
\textbf{Score Level} & \textbf{Evaluation Criteria (Helpfulness)} \\
\midrule
1 -- Obstacle & Fails to parse data or produces non-executable code; requires total restart. \\
2 -- Low Utility & Provides shallow insights; requires significant manual correction. \\
3 -- Average & Automates routine tasks but lacks depth; needs moderate human guidance. \\
4 -- Good & Understands data well; generates executable scripts with only minor gaps. \\
5 -- Excellent & Acts as an autonomous collaborator; provides deep insights and ready-to-use code. \\
\bottomrule
\end{tabularx}
\label{tab:human_rubric}
\end{table}

\subsection{Case Study}

\paragraph{End-to-end Research}

We present an end-to-end workflow where \sciagent autonomously investigates calibrated few-shot knowledge tracing in cold-start educational settings. After analyzing datasets like ASSISTments~\cite{feng2009addressing}, the system proposed a novel direction: student-conditioned evidential meta-learning with hyper-conditioned priors. \sciagent executed the entire research pipeline from method design and model implementation to quantitative evaluation. The generated paper assesses the proposed ``EvidentialHyperMAML'' against baselines using classic metrics (e.g., AUC-ROC). Fig.~\ref{fig:scider-evidence} highlights the comprehensive generated manuscript, detailing the methodology, empirical results, and limitations.

\paragraph{Visualization}
To demonstrate our framework's domain-specific visual reasoning, we present an AstroVisBench qualitative case study. Astronomical visualization is highly out-of-distribution, often requiring complex coordinate transformations and specialized APIs. \sciagent addresses this through an iterative multimodal feedback loop: it generates an initial visualization script from raw data, acts as a visual critic to inspect the rendered plot for anomalies (e.g., incorrect spectral scaling or misaligned celestial axes), and autonomously refines the code.
Fig.~\ref{fig:astrovis_examples} demonstrates how our self-correcting multimodal agent generates high-fidelity astronomical figures, successfully overcoming the geometric and formatting errors common in standard baseline models.


\section{Conclusion}
We present \sciagent, an autonomous, data-centric system that automates the scientific research lifecycle. To overcome the limitations of current agents in processing raw, domain-specific data, our framework integrates a dynamic multimodal skill system across four specialized agents: Ideation, Data Analysis, Experimentation, and Critic. By grounding abstract reasoning in rigorous data analysis, \sciagent effectively bridges the gap between high-level hypothesis generation and reproducible code synthesis.

To democratize AI-driven research, we release the \opensci-SFT-8K trajectory dataset and the fine-tuned \opensci-27B model. Evaluations demonstrate our system's promising performance in idea generation, complex machine learning, and multidisciplinary multimodal reasoning. 
Distributed as a modular Python package, \sciagent is highly accessible, enabling researchers to easily initiate closed-loop workflows. This work lowers technical barriers, accelerates interdisciplinary discovery, and promotes the widespread adoption of autonomous, end-to-end research partners.

\bibliographystyle{IEEEtran} 
\bibliography{custom}

@misc{lu2024ai,
  title         = {The {AI} Scientist: Towards Fully Automated Open-Ended Scientific Discovery},
  author        = {Lu, Chris and Lu, Cong and Lange, Robert Tjarko and Foerster, Jakob and Clune, Jeff and Ha, David},
  year          = {2024},
  eprint        = {2408.06292},
  archivePrefix = {arXiv},
  primaryClass  = {cs.AI},
  doi           = {10.48550/arXiv.2408.06292},
  url           = {https://arxiv.org/abs/2408.06292}
}

@misc{yamada2025ai,
  title         = {The {AI} Scientist-v2: Workshop-Level Automated Scientific Discovery via Agentic Tree Search},
  author        = {Yamada, Yutaro and Lange, Robert Tjarko and Lu, Cong and Hu, Shengran and Lu, Chris and Foerster, Jakob and Clune, Jeff and Ha, David},
  year          = {2025},
  eprint        = {2504.08066},
  archivePrefix = {arXiv},
  primaryClass  = {cs.AI},
  doi           = {10.48550/arXiv.2504.08066},
  url           = {https://arxiv.org/abs/2504.08066}
}

@inproceedings{tang2025ai,
  title         = {{AI}-Researcher: Autonomous Scientific Innovation},
  author        = {Tang, Jiabin and Xia, Lianghao and Li, Zhonghang and Huang, Chao},
  booktitle     = {Advances in Neural Information Processing Systems},
  volume        = {38},
  year          = {2025},
  url           = {https://openreview.net/forum?id=kQWyOYUAC4},
  note          = {NeurIPS 2025 spotlight},
  eprint        = {2505.18705},
  archivePrefix = {arXiv},
  primaryClass  = {cs.AI},
  doi           = {10.48550/arXiv.2505.18705}
}

@inproceedings{schmidgall2025agent,
  title     = {Agent Laboratory: Using {LLM} Agents as Research Assistants},
  author    = {Schmidgall, Samuel and Su, Yusheng and Wang, Ze and Sun, Ximeng and Wu, Jialian and Yu, Xiaodong and Liu, Jiang and Moor, Michael and Liu, Zicheng and Barsoum, Emad},
  editor    = {Christodoulopoulos, Christos and Chakraborty, Tanmoy and Rose, Carolyn and Peng, Violet},
  booktitle = {Findings of the Association for Computational Linguistics: EMNLP 2025},
  month     = nov,
  year      = {2025},
  address   = {Suzhou, China},
  publisher = {Association for Computational Linguistics},
  url       = {https://aclanthology.org/2025.findings-emnlp.320/},
  doi       = {10.18653/v1/2025.findings-emnlp.320},
  pages     = {5977--6043},
  ISBN      = {979-8-89176-335-7}
}

@inproceedings{yu2025tinyscientist,
  title     = {{T}iny{S}cientist: An Interactive, Extensible, and Controllable Framework for Building Research Agents},
  author    = {Yu, Haofei and Xuan, Keyang and Li, Fenghai and Zhu, Kunlun and Lei, Zijie and Zhang, Jiaxun and Qi, Ziheng and Richardson, Kyle and You, Jiaxuan},
  editor    = {Habernal, Ivan and Schulam, Peter and Tiedemann, J{\"o}rg},
  booktitle = {Proceedings of the 2025 Conference on Empirical Methods in Natural Language Processing: System Demonstrations},
  month     = nov,
  year      = {2025},
  address   = {Suzhou, China},
  publisher = {Association for Computational Linguistics},
  url       = {https://aclanthology.org/2025.emnlp-demos.41/},
  doi       = {10.18653/v1/2025.emnlp-demos.41},
  pages     = {558--590},
  ISBN      = {979-8-89176-334-0}
}

@misc{zhang2025deepanalyze,
  title         = {{DeepAnalyze}: Agentic Large Language Models for Autonomous Data Science},
  author        = {Zhang, Shaolei and Fan, Ju and Fan, Meihao and Li, Guoliang and Du, Xiaoyong},
  year          = {2025},
  eprint        = {2510.16872},
  archivePrefix = {arXiv},
  primaryClass  = {cs.AI},
  doi           = {10.48550/arXiv.2510.16872},
  url           = {https://arxiv.org/abs/2510.16872}
}

@misc{team2025internagent,
  title         = {InternAgent: When Agent Becomes the Scientist--Building Closed-Loop System from Hypothesis to Verification},
  author        = {InternAgent Team and Bo Zhang and Shiyang Feng and Xiangchao Yan and Jiakang Yuan and Runmin Ma and Yusong Hu and others},
  year          = {2025},
  eprint        = {2505.16938},
  archivePrefix = {arXiv},
  primaryClass  = {cs.AI},
  doi           = {10.48550/arXiv.2505.16938},
  url           = {https://arxiv.org/abs/2505.16938}
}

@misc{feng2026internagent,
  title         = {Internagent-1.5: A unified agentic framework for long-horizon autonomous scientific discovery},
  author        = {Shiyang Feng and Runmin Ma and Xiangchao Yan and Yue Fan and Yusong Hu and Songtao Huang and Shuaiyu Zhang and others},
  year          = {2026},
  eprint        = {2602.08990},
  archivePrefix = {arXiv},
  primaryClass  = {cs.AI},
  doi           = {10.48550/arXiv.2602.08990},
  url           = {https://arxiv.org/abs/2602.08990}
}

@misc{qiu2025ai,
  title         = {{AI} Idea Bench 2025: {AI} Research Idea Generation Benchmark},
  author        = {Qiu, Yansheng and Zhang, Haoquan and Xu, Zhaopan and Li, Ming and Song, Diping and Wang, Zheng and Zhang, Kaipeng},
  year          = {2025},
  eprint        = {2504.14191},
  archivePrefix = {arXiv},
  primaryClass  = {cs.AI},
  doi           = {10.48550/arXiv.2504.14191},
  url           = {https://arxiv.org/abs/2504.14191}
}

@inproceedings{chan2025mle,
  title         = {{MLE-Bench}: Evaluating Machine Learning Agents on Machine Learning Engineering},
  author        = {Chan, Jun Shern and Chowdhury, Neil and Jaffe, Oliver and Aung, James and Sherburn, Dane and Mays, Evan and Starace, Giulio and Liu, Kevin and Maksin, Leon and Patwardhan, Tejal and Madry, Aleksander and Weng, Lilian},
  booktitle     = {International Conference on Learning Representations},
  year          = {2025},
  url           = {https://openreview.net/forum?id=6s5uXNWGIh},
  eprint        = {2410.07095},
  archivePrefix = {arXiv},
  primaryClass  = {cs.AI},
  doi           = {10.48550/arXiv.2410.07095}
}

@misc{jiang2025aide,
  title         = {{AIDE}: {AI}-Driven Exploration in the Space of Code},
  author        = {Jiang, Zhengyao and Schmidt, Dominik and Srikanth, Dhruv and Xu, Dixing and Kaplan, Ian and Jacenko, Deniss and Wu, Yuxiang},
  year          = {2025},
  eprint        = {2502.13138},
  archivePrefix = {arXiv},
  primaryClass  = {cs.AI},
  doi           = {10.48550/arXiv.2502.13138},
  url           = {https://arxiv.org/abs/2502.13138}
}

@inproceedings{toledo2025aira,
  title         = {{AI} Research Agents for Machine Learning: Search, Exploration, and Generalization in {MLE-bench}},
  author        = {Toledo, Edan and Hambardzumyan, Karen and Josifoski, Martin and Hazra, Rishi and Baldwin, Nicolas and Audran-Reiss, Alexis and Kuchnik, Michael and others},
  booktitle     = {Advances in Neural Information Processing Systems},
  volume        = {38},
  year          = {2025},
  url           = {https://proceedings.neurips.cc/paper_files/paper/2025/hash/328b81881da145412f2bc56c998dfb6a-Abstract-Conference.html},
  eprint        = {2507.02554},
  archivePrefix = {arXiv},
  primaryClass  = {cs.AI},
  doi           = {10.48550/arXiv.2507.02554}
}

@misc{liu2025mlmaster,
  title         = {{ML-Master}: Towards {AI}-for-{AI} via Integration of Exploration and Reasoning},
  author        = {Liu, Zexi and Cai, Yuzhu and Zhu, Xinyu and Zheng, Yujie and Chen, Runkun and Wen, Ying and Wang, Yanfeng and E, Weinan and Chen, Siheng},
  year          = {2025},
  eprint        = {2506.16499},
  archivePrefix = {arXiv},
  primaryClass  = {cs.AI},
  doi           = {10.48550/arXiv.2506.16499},
  url           = {https://arxiv.org/abs/2506.16499}
}

@inproceedings{tian2024scicode,
  author = {Tian, Minyang and Gao, Luyu and Zhang, Shizhuo Dylan and Chen, Xinan and Fan, Cunwei and Guo, Xuefei and Haas, Roland and others},
  title = {SciCode: a research coding benchmark curated by scientists},
  year = {2024},
  isbn = {9798331314385},
  publisher = {Curran Associates Inc.},
  address = {Red Hook, NY, USA},
  booktitle = {Proceedings of the 38th International Conference on Neural Information Processing Systems},
  articleno = {963},
  numpages = {27},
  location = {Vancouver, BC, Canada},
  series = {NIPS '24}
}

@inproceedings{su2025many,
  title     = {Many Heads Are Better Than One: Improved Scientific Idea Generation by A {LLM}-Based Multi-Agent System},
  author    = {Su, Haoyang and Chen, Renqi and Tang, Shixiang and Yin, Zhenfei and Zheng, Xinzhe and Li, Jinzhe and others},
  booktitle = {Proceedings of the 63rd Annual Meeting of the Association for Computational Linguistics (Volume 1: Long Papers)},
  month     = jul,
  year      = {2025},
  address   = {Vienna, Austria},
  publisher = {Association for Computational Linguistics},
  url       = {https://aclanthology.org/2025.acl-long.1368/},
  doi       = {10.18653/v1/2025.acl-long.1368},
  pages     = {28201--28240},
  ISBN      = {979-8-89176-251-0}
}

@misc{wang2024scipip,
  title         = {{SciPIP}: An {LLM}-Based Scientific Paper Idea Proposer},
  author        = {Wang, Wenxiao and Gu, Lihui Ribbon and Zhang, Liye and Luo, Yunxiang and Dai, Yi and Shen, Chen and Xie, Liang and Lin, Binbin and He, Xiaofei and Ye, Jieping},
  year          = {2024},
  eprint        = {2410.23166},
  archivePrefix = {arXiv},
  primaryClass  = {cs.CL},
  doi           = {10.48550/arXiv.2410.23166},
  url           = {https://arxiv.org/abs/2410.23166}
}

@misc{arxiv,
  title        = {arXiv.org e-Print Archive},
  author       = {{arXiv}},
  year         = {2026},
  howpublished = {\url{https://arxiv.org}},
  url          = {https://arxiv.org},
  urldate      = {2026-05-23},
  note         = {Accessed: 2026-05-23}
}

@misc{semanticscholar,
  title        = {Semantic Scholar},
  author       = {{Allen Institute for AI}},
  year         = {2026},
  howpublished = {\url{https://www.semanticscholar.org}},
  url          = {https://www.semanticscholar.org},
  urldate      = {2026-05-23},
  note         = {Accessed: 2026-05-23}
}

@misc{pubmed,
  title        = {PubMed},
  author       = {{National Center for Biotechnology Information}},
  year         = {2026},
  howpublished = {\url{https://pubmed.ncbi.nlm.nih.gov}},
  url          = {https://pubmed.ncbi.nlm.nih.gov},
  urldate      = {2026-05-23},
  note         = {Accessed: 2026-05-23}
}

@misc{luo2026benchmarkingaiscientistsomics,
      title={Benchmarking AI scientists for omics data driven biological discovery}, 
      author={Erpai Luo and Jinmeng Jia and Yifan Xiong and Xiangyu Li and Xiaobo Guo and Baoqi Yu and Minsheng Hao and Lei Wei and Xuegong Zhang},
      year={2026},
      eprint={2505.08341},
      archivePrefix={arXiv},
      primaryClass={cs.AI},
      url={https://arxiv.org/abs/2505.08341}, 
}

@Article{white2023large,
  author ="White, Andrew D. and Hocky, Glen M. and Gandhi, Heta A. and Ansari, Mehrad and Cox, Sam and Wellawatte, Geemi P. and Sasmal, Subarna and Yang, Ziyue and Liu, Kangxin and Singh, Yuvraj and Peña Ccoa, Willmor J.",
  title  ="Assessment of chemistry knowledge in large language models that generate code",
  journal  ="Digital Discovery",
  year  ="2023",
  volume  ="2",
  issue  ="2",
  pages  ="368-376",
  publisher  ="RSC",
  doi  ="10.1039/D2DD00087C",
  url  ="http://dx.doi.org/10.1039/D2DD00087C"
}

@inproceedings{starace2025paperbenchevaluatingaisability,
  title     = {{P}aper{B}ench: Evaluating {AI}'s Ability to Replicate {AI} Research},
  author    = {Starace, Giulio and Jaffe, Oliver and Sherburn, Dane and Aung, James and Chan, Jun Shern and Maksin, Leon and others},
  booktitle = {Proceedings of the 42nd International Conference on Machine Learning},
  pages     = {56843--56873},
  year      = {2025},
  editor    = {Singh, Aarti and Fazel, Maryam and Hsu, Daniel and Lacoste-Julien, Simon and Berkenkamp, Felix and Maharaj, Tegan and Wagstaff, Kiri and Zhu, Jerry},
  volume    = {267},
  series    = {Proceedings of Machine Learning Research},
  month     = {13--19 Jul},
  publisher = {PMLR},
  url       = {https://proceedings.mlr.press/v267/starace25a.html},
  eprint    = {2504.01848},
  archivePrefix = {arXiv},
  primaryClass  = {cs.AI}
}

@article{Wang2024GeneAgentSL,
  title     = {{GeneAgent}: Self-Verification Language Agent for Gene-Set Analysis Using Domain Databases},
  author    = {Wang, Zhizheng and Jin, Qiao and Wei, Chih-Hsuan and Tian, Shubo and Lai, Po-Ting and Zhu, Qingqing and Day, Chi-Ping and Ross, Christina and Leaman, Robert and Lu, Zhiyong},
  journal   = {Nature Methods},
  volume    = {22},
  pages     = {1677--1685},
  year      = {2025},
  doi       = {10.1038/s41592-025-02748-6},
  url       = {https://doi.org/10.1038/s41592-025-02748-6}
}

@misc{gottweis2025towards,
  title         = {Towards an {AI} Co-Scientist},
  author        = {Gottweis, Juraj and Weng, Wei-Hung and Daryin, Alexander and Tu, Tao and Palepu, Anil and Sirkovic, Petar and others},
  year          = {2025},
  eprint        = {2502.18864},
  archivePrefix = {arXiv},
  primaryClass  = {cs.AI},
  doi           = {10.48550/arXiv.2502.18864},
  url           = {https://arxiv.org/abs/2502.18864}
}

@inproceedings{garikaparthi-etal-2025-iris,
  title     = {{IRIS}: Interactive Research Ideation System for Accelerating Scientific Discovery},
  author    = {Garikaparthi, Aniketh and Patwardhan, Manasi and Vig, Lovekesh and Cohan, Arman},
  editor    = {Mishra, Pushkar and Muresan, Smaranda and Yu, Tao},
  booktitle = {Proceedings of the 63rd Annual Meeting of the Association for Computational Linguistics (Volume 3: System Demonstrations)},
  month     = jul,
  year      = {2025},
  address   = {Vienna, Austria},
  publisher = {Association for Computational Linguistics},
  url       = {https://aclanthology.org/2025.acl-demo.57/},
  doi       = {10.18653/v1/2025.acl-demo.57},
  pages     = {592--603},
  ISBN      = {979-8-89176-253-4}
}

@misc{comanici2025gemini,
  title         = {{Gemini 2.5}: Pushing the Frontier with Advanced Reasoning, Multimodality, Long Context, and Next Generation Agentic Capabilities},
  author        = {Comanici, Gheorghe and Bieber, Eric and Schaekermann, Mike and Pasupat, Ice and Sachdeva, Noveen and Dhillon, Inderjit and Blistein, Marcel and Ram, Ori and Zhang, Dan and Rosen, Evan and others},
  year          = {2025},
  eprint        = {2507.06261},
  archivePrefix = {arXiv},
  primaryClass  = {cs.CL},
  doi           = {10.48550/arXiv.2507.06261},
  url           = {https://arxiv.org/abs/2507.06261}
}

@techreport{anthropic2025claude4systemcard,
  author      = {{Anthropic}},
  title       = {System Card: Claude Opus 4 \& Claude Sonnet 4},
  institution = {Anthropic},
  year        = {2025},
  month       = may,
  url         = {https://www-cdn.anthropic.com/6d8a8055020700718b0c49369f60816ba2a7c285.pdf},
  urldate     = {2026-05-23},
  note        = {Accessed: 2026-05-23}
}

@misc{song2026paperorchestra,
      title={PaperOrchestra: A Multi-Agent Framework for Automated AI Research Paper Writing}, 
      author={Yiwen Song and Yale Song and Tomas Pfister and Jinsung Yoon},
      year={2026},
      eprint={2604.05018},
      archivePrefix={arXiv},
      primaryClass={cs.AI},
      url={https://arxiv.org/abs/2604.05018}, 
}

@inproceedings{majumder2024discoverybench,
  title     = {{DiscoveryBench}: Towards Data-Driven Discovery with Large Language Models},
  author    = {Majumder, Bodhisattwa Prasad and Surana, Harshit and Agarwal, Dhruv and Mishra, Bhavana Dalvi and Meena, Abhijeetsingh and Prakhar, Aryan and Vora, Tirth and Khot, Tushar and Sabharwal, Ashish and Clark, Peter},
  booktitle = {International Conference on Learning Representations},
  year      = {2025},
  url       = {https://openreview.net/forum?id=vyflgpwfJW},
  eprint    = {2407.01725},
  archivePrefix = {arXiv},
  primaryClass  = {cs.AI},
  doi       = {10.48550/arXiv.2407.01725}
}

@misc{lupidi2026airsbench,
  title         = {{AIRS-Bench}: A Suite of Tasks for Frontier {AI} Research Science Agents},
  author        = {Lupidi, Alisia and Gauri, Bhavul and Foster, Thomas Simon and Al~Omari, Bassel and Magka, Despoina and Pepe, Alberto and Audran-Reiss, Alexis and others},
  year          = {2026},
  eprint        = {2602.06855},
  archivePrefix = {arXiv},
  primaryClass  = {cs.AI},
  doi           = {10.48550/arXiv.2602.06855},
  url           = {https://arxiv.org/abs/2602.06855}
}

@inproceedings{joseph2025astrovisbench,
  title         = {{AstroVisBench}: A Code Benchmark for Scientific Computing and Visualization in Astronomy},
  author        = {Joseph, Sebastian and Husain, Syed M. and Offner, Stella and Juneau, St{\'e}phanie and Torrey, Paul and Bolton, Adam and Farias, Juan and Gaffney, Niall and Durrett, Greg and Li, Junyi Jessy},
  booktitle     = {Advances in Neural Information Processing Systems},
  volume        = {38},
  year          = {2025},
  url           = {https://proceedings.neurips.cc/paper_files/paper/2025/hash/c2d9a98d55e7f1e4e88dd15d900c9c17-Abstract-Datasets_and_Benchmarks_Track.html},
  eprint        = {2505.20538},
  archivePrefix = {arXiv},
  primaryClass  = {cs.AI},
  doi           = {10.48550/arXiv.2505.20538}
}

@inproceedings{yao2022react,
  title         = {React: Synergizing reasoning and acting in language models},
  author        = {Yao, Shunyu and Zhao, Jeffrey and Yu, Dian and Du, Nan and Shafran, Izhak and Narasimhan, Karthik and Cao, Yuan},
  booktitle     = {International Conference on Learning Representations},
  year          = {2022},
  eprint        = {2210.03629},
  archivePrefix = {arXiv},
  primaryClass  = {cs.CL},
  doi           = {10.48550/arXiv.2210.03629}
}

@misc{zhang2025datascibench,
  title         = {Datascibench: An llm agent benchmark for data science},
  author        = {Zhang, Dan and Zhoubian, Sining and Cai, Min and Li, Fengzu and Yang, Lekang and Wang, Wei and Dong, Tianjiao and Hu, Ziniu and Tang, Jie and Yue, Yisong},
  year          = {2025},
  eprint        = {2502.13897},
  archivePrefix = {arXiv},
  primaryClass  = {cs.CL},
  doi           = {10.48550/arXiv.2502.13897},
  url           = {https://arxiv.org/abs/2502.13897}
}

@inproceedings{lai2023ds,
  title={DS-1000: A natural and reliable benchmark for data science code generation},
  author={Lai, Yuhang and Li, Chengxi and Wang, Yiming and Zhang, Tianyi and Zhong, Ruiqi and Zettlemoyer, Luke and Yih, Wen-tau and Fried, Daniel and Wang, Sida and Yu, Tao},
  booktitle={International Conference on Machine Learning},
  pages={18319--18345},
  year={2023},
  organization={PMLR},
  doi={10.48550/arXiv.2211.11501}
}

@inproceedings{jing2024dsbench,
  title         = {DSBench: How Far Are Data Science Agents from Becoming Data Science Experts?},
  author        = {Jing, Liqiang and Huang, Zhehui and Wang, Xiaoyang and Yao, Wenlin and Yu, Wenhao and Ma, Kaixin and Zhang, Hongming and Du, Xinya and Yu, Dong},
  booktitle     = {International Conference on Learning Representations},
  year          = {2025},
  eprint        = {2409.07703},
  archivePrefix = {arXiv},
  primaryClass  = {cs.AI},
  doi           = {10.48550/arXiv.2409.07703}
}

@inproceedings{chen2025scienceagentbench,
  title         = {ScienceAgentBench: Toward Rigorous Assessment of Language Agents for Data-Driven Scientific Discovery},
  author        = {Ziru Chen and Shijie Chen and Yuting Ning and Qianheng Zhang and Boshi Wang and Botao Yu and Yifei Li and others},
  booktitle     = {International Conference on Learning Representations},
  year          = {2025},
  eprint        = {2410.05080},
  archivePrefix = {arXiv},
  primaryClass  = {cs.CL},
  doi           = {10.48550/arXiv.2410.05080}
}

@misc{majumder2024data,
  title         = {Data-driven discovery with large generative models},
  author        = {Majumder, Bodhisattwa Prasad and Surana, Harshit and Agarwal, Dhruv and Hazra, Sanchaita and Sabharwal, Ashish and Clark, Peter},
  year          = {2024},
  eprint        = {2402.13610},
  archivePrefix = {arXiv},
  primaryClass  = {cs.LG},
  doi           = {10.48550/arXiv.2402.13610},
  url           = {https://arxiv.org/abs/2402.13610}
}

@misc{allenai2026asta,
  title        = {Asta Agents: {AI} Tools for Scientific Research},
  author       = {{Allen Institute for AI}},
  year         = {2026},
  howpublished = {\url{https://allenai.org/asta/agents}},
  url          = {https://allenai.org/asta/agents},
  urldate      = {2026-05-24},
  note         = {Accessed: 2026-05-24}
}

@inproceedings{hu2022lora,
  title         = {LoRA: Low-Rank Adaptation of Large Language Models},
  author        = {Hu, Edward J and Shen, Yelong and Wallis, Phillip and Allen-Zhu, Zeyuan and Li, Yuanzhi and Wang, Shean and Wang, Lu and Chen, Weizhu},
  booktitle     = {International Conference on Learning Representations},
  year          = {2022},
  eprint        = {2106.09685},
  archivePrefix = {arXiv},
  primaryClass  = {cs.CL},
  doi           = {10.48550/arXiv.2106.09685}
}

@inproceedings{rajbhandari2020zero,
  title={Zero: Memory optimizations toward training trillion parameter models},
  author={Rajbhandari, Samyam and Rasley, Jeff and Ruwase, Olatunji and He, Yuxiong},
  booktitle={SC20: international conference for high performance computing, networking, storage and analysis},
  pages={1--16},
  year={2020},
  organization={IEEE},
  doi={10.1109/SC41405.2020.00024}
}

@inproceedings{peng2024yarn,
  title         = {Yarn: Efficient context window extension of large language models},
  author        = {Peng, Bowen and Quesnelle, Jeffrey and Fan, Honglu and Shippole, Enrico},
  booktitle     = {International Conference on Learning Representations},
  year          = {2024},
  eprint        = {2309.00071},
  archivePrefix = {arXiv},
  primaryClass  = {cs.CL},
  doi           = {10.48550/arXiv.2309.00071}
}

@inproceedings{chen2024selfdebug,
  title         = {Teaching Large Language Models to Self-Debug},
  author        = {Chen, Xinyun and Lin, Maxwell and Sch{\"a}rli, Nathanael and Zhou, Denny},
  booktitle     = {International Conference on Learning Representations},
  year          = {2024},
  eprint        = {2304.05128},
  archivePrefix = {arXiv},
  primaryClass  = {cs.CL},
  doi           = {10.48550/arXiv.2304.05128}
}

@inproceedings{qin2024toolllm,
  title         = {{ToolLLM}: Facilitating Large Language Models to Master 16000+ Real-world {APIs}},
  author        = {Qin, Yujia and Liang, Shihao and Ye, Yining and Zhu, Kunlun and Yan, Lan and Lu, Yaxi and Lin, Yankai and Cong, Xin and Tang, Xiangru and Qian, Bill and others},
  booktitle     = {International Conference on Learning Representations},
  year          = {2024},
  eprint        = {2307.16789},
  archivePrefix = {arXiv},
  primaryClass  = {cs.AI},
  doi           = {10.48550/arXiv.2307.16789}
}

@inproceedings{zheng2023judging,
  author = {Zheng, Lianmin and Chiang, Wei-Lin and Sheng, Ying and Zhuang, Siyuan and Wu, Zhanghao and Zhuang, Yonghao and others},
  title = {Judging LLM-as-a-judge with MT-bench and Chatbot Arena},
  year = {2023},
  publisher = {Curran Associates Inc.},
  address = {Red Hook, NY, USA},
  booktitle = {Proceedings of the 37th International Conference on Neural Information Processing Systems},
  articleno = {2020},
  numpages = {29},
  location = {New Orleans, LA, USA},
  series = {NIPS '23}
}

@techreport{gdm2025gemini3,
  author      = {{Google DeepMind}},
  title       = {Gemini 3 Pro Model Card},
  institution = {Google DeepMind},
  year        = {2025},
  month       = nov,
  url         = {https://storage.googleapis.com/deepmind-media/Model-Cards/Gemini-3-Pro-Model-Card.pdf},
  note        = {Accessed: 2026-06-03}
}

@misc{qwen2026qwen36,
  author       = {{Qwen Team}},
  title        = {Qwen3.6},
  year         = {2026},
  month        = apr,
  howpublished = {\url{https://github.com/QwenLM/Qwen3.6}},
  note         = {Open-weight Qwen3.6-27B; accessed: 2026-06-03}
}

@article{feng2009addressing,
  title={Addressing the assessment challenge with an online system that tutors as it assesses},
  author={Feng, Mingyu and Heffernan, Neil and Koedinger, Kenneth},
  journal={User modeling and user-adapted interaction},
  volume={19},
  number={3},
  pages={243--266},
  year={2009},
  publisher={Springer},
  url={https://doi.org/10.1007/s11257-009-9063-7},
  doi={10.1007/s11257-009-9063-7}
}

\appendix

\begin{figure}[htbp]
\centering
\includegraphics[width=0.9\linewidth]{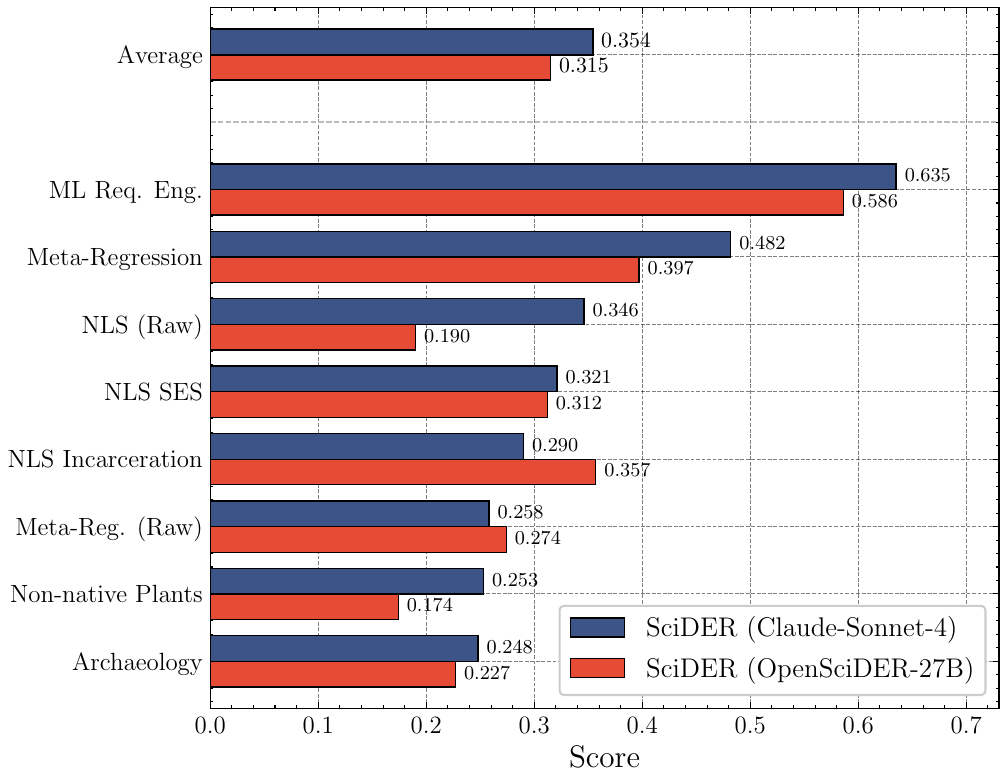}
\caption{
Per-task scores of our method on DiscoveryBench.
}
\label{fig:discoverybench}
\end{figure}

\begin{figure}[htbp]
\centering
\includegraphics[width=0.95\linewidth]{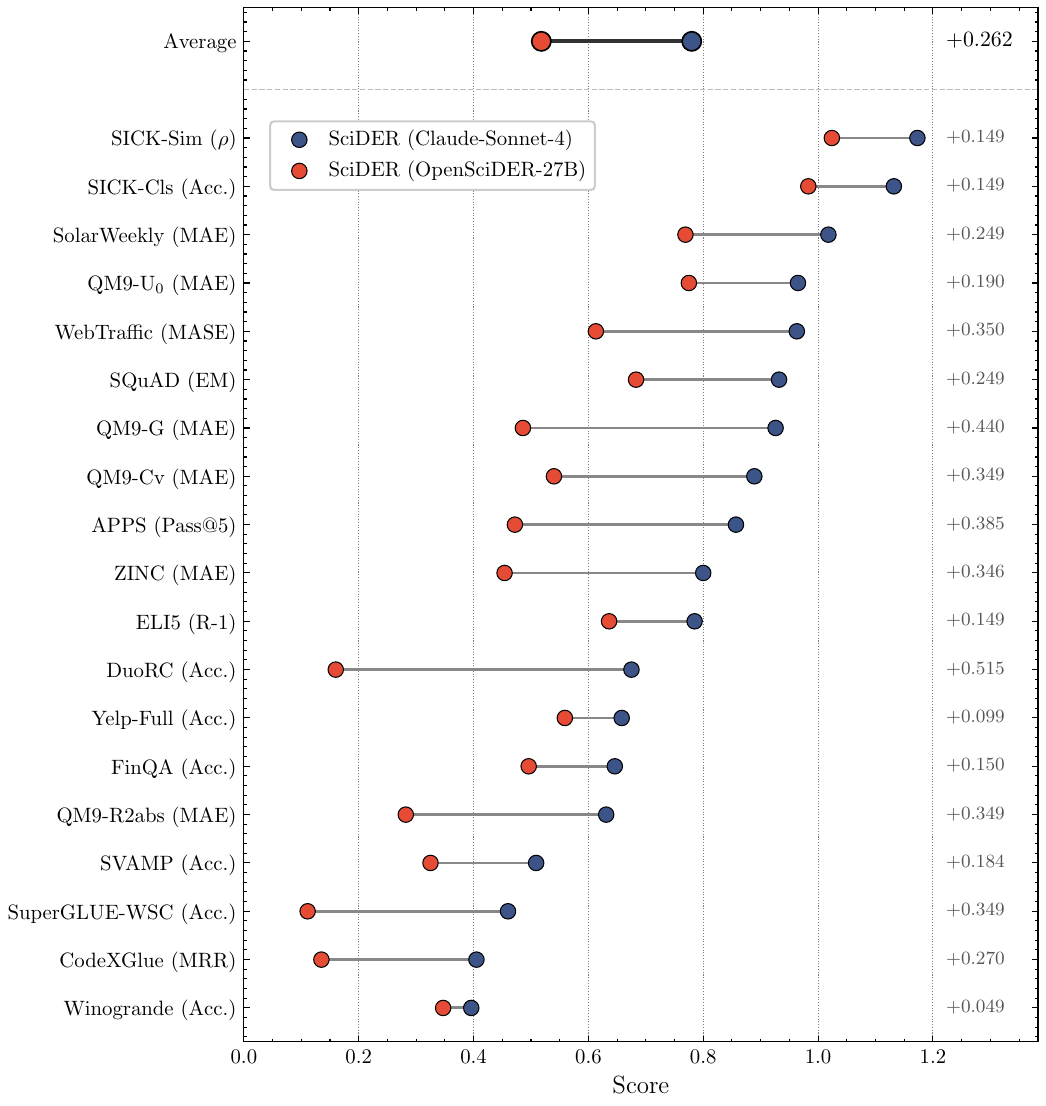}
\caption{
Per-task scores of our method on AIRS-Bench.
}
\label{fig:airsbench_pertask}
\end{figure}

\end{document}